\documentclass[acmsmall]{acmart}

\AtBeginDocument{%
  \providecommand\BibTeX{{%
    \normalfont B\kern-0.5em{\scshape i\kern-0.25em b}\kern-0.8em\TeX}}}

\usepackage{multirow}
\usepackage{graphicx}
\usepackage{booktabs}
\usepackage{amsmath}
\usepackage{amsfonts}
\DeclareMathOperator*{\argmax}{arg\,max}

\begin{document}

\title{A Survey of Natural Language Generation}

\author{Chenhe Dong}
\authornote{Equal contribution}
\email{dongchh@mail2.sysu.edu.cn}
\affiliation{
  \institution{Sun Yat-Sen University}
  \city{Guangzhou}
  \country{China}
  \postcode{510275}}

\author{Yinghui Li}
\authornotemark[1]
\email{liyinghu20@mails.tsinghua.edu.cn}
\affiliation{
  \institution{Tsinghua University}
  \city{Shenzhen}
  \country{China}
  \postcode{518055}}

\author{Haifan Gong}
\email{gonghf@mail2.sysu.edu.cn}
\affiliation{
  \institution{Sun Yat-Sen University}
  \city{Guangzhou}
  \country{China}
  \postcode{510275}}

\author{Miaoxin Chen}
\email{cmx20@mails.tsinghua.edu.cn}
\affiliation{
  \institution{Tsinghua University}
  \city{Shenzhen}
  \country{China}
  \postcode{518055}}
  
\author{Junxin Li}
\email{ljx20@mails.tsinghua.edu.cn}
  \affiliation{\institution{Tsinghua University}
  \city{Shenzhen}
  \country{China}
  \postcode{518055}}

\author{Ying Shen}
\authornote{Corresponding author}
\email{sheny76@mail.sysu.edu.cn}
\affiliation{
  \institution{Sun Yat-Sen University}
  \city{Guangzhou}
  \country{China}
  \postcode{510275}}

\author{Min Yang}
\authornotemark[2]
\email{min.yang@siat.ac.cn}
\affiliation{
  \institution{Chinese Academy of Science}
  \city{Shenzhen}
  \country{China}
  \postcode{510100}}

\renewcommand{\shortauthors}{Dong and Li, et al.}

\begin{abstract}
This paper offers a comprehensive review of the research on Natural Language Generation (NLG) over the past two decades, especially in relation to data-to-text generation and text-to-text generation deep learning methods, as well as new applications of NLG technology.
This survey aims to (a) give the latest synthesis of deep learning research on the NLG core tasks, as well as the architectures adopted in the field; (b) detail meticulously and comprehensively various NLG tasks and datasets, and draw attention to the challenges in NLG evaluation, focusing on different evaluation methods and their relationships; (c) highlight some future emphasis and relatively recent research issues that arise due to the increasing synergy between NLG and other artificial intelligence areas, such as computer vision, text and computational creativity.
\end{abstract}



\setcopyright{acmcopyright}
\acmJournal{CSUR}
\acmYear{2022} \acmVolume{1} \acmNumber{1} \acmArticle{1} \acmMonth{1} \acmPrice{15.00}\acmDOI{10.1145/3554727}

\maketitle
\section{Introduction}
\label{sec:introduction}

This paper surveys the current state of the art in Natural Language Generation (NLG), defined as the task of generating text from underlying non-linguistic representation of information \cite{reiter1997building}.
NLG has been receiving more and more attention from researchers because of its extremely challenging and promising application prospects.

\subsection{What is Natural Language Generation?}
Natural Language Generation (NLG) is the process of producing a natural language text in order to meet specified communicative goals. The texts that are generated may range from a single phrase given in answer to a question, through multi-sentence remarks and questions within a dialog, to full-page explanations.

In contrast with the organization of the Natural Language Understanding (NLU) process -- which can follow the traditional stages of a linguistic analysis: morphology, syntax, semantics, pragmatics/discourse -- the generation process has a fundamentally different character.
Generation proceeds involve the content planning, determination and realization from content to form, from intentions and perspectives to linearly arrayed words and syntactic markers. Coupled with its application, the situation, and the discourse, they provide the basis for making choices among the alternative wordings and constructions that the language provides, which is the primary effort in constructing a text deliberately \cite{mcdonald2010natural}. With its opposite flow of information, one might assume that a generation process could be organized like an understanding process but with the stages in opposite order. 

Both data-to-text generation and text-to-text generation are instances of NLG.
Generating text from image is an application of data-to-text generation.
A further text-to-text generation complication is dividing NLG tasks into three categories, i.e., text abbreviation, text expansion, text rewriting and reasoning. 
The text abbreviation task is formulated to condense information from long texts to short ones, typically including research on text summarization \cite{song2019mass,lewis2020bart,cai2019improving,qi2020prophetnet,ding2020generative,cao2020factual,dong2020multi}, question generation \cite{dw2021question,bhatia2019dynamic,serban2016generating,yuan2017machine,zhao2018paragraph-level,sun2018answer,kim2019improving,wang2019a-multi-agent,jia2020how,liu2019learning,wang2020pathqg}, and distractor generation \cite{stasaski2017multiple,liang2018distractor,gao2019generating,qiu2020automatic,maurya2020learning,srinivas2019towards,ren2021knowledge-driven,patra2019a-hybrid}. 
The text expansion tasks, such as short text expansion \cite{billerbeck2003query,tang2017end-to-end,shi2019functional,safovish2020fiction} and topic-to-essay generation \cite{feng2018topic-to-essay,wang2020self-attention,yuan2019efficient,yang2019enhancing,qiao2020a-sentiment-controllable}, generate complete sentences or even texts from some meaningful words by considering and adding elements like conjunctions and prepositions to transform the input words into linguistically correct outputs.
The goal of text rewriting and reasoning task is to rewrite the text into another style or applying reasoning methods to create responses. There are two sub-tasks: text style transfer \cite{jhamtani2017shakespearizing,zhao2018adversarially,chen2018adversarial,xu2018unpaired,santos2018fighting,prabhumoye2018style,luo2019a-dual,fu2018style,mir2019evaluating,lin2020data}, and dialogue generation \cite{zhou2017mechanism-aware,xu2017neural,luo2018an-auto-encoder,li2020attention,lian2019learning,jiang2020knowledge,wang2020knowledge,bao2020plato}. 
The task of visual based text generation targets at generate the explanation or summarization of the given image or video, involving the study of image caption \cite{vinyals2015show,anderson2018bottom-up,lu2017knowing,yao2018exploring,rennie2017self-critical,yang2019auto-encoding}, video caption \cite{xu2016msrvtt, krishna2017activitynet, sigurdsson2016charades, sun2019videobert,donahue2015long,venugopalan2016improving,wu2019densely,zhou2018end-to-end,wang2018a-reinforced,park2019adversarial,zhu2020actbert,lei2021less}, and visual storytelling \cite{lin2019informative,yang2019knowledgeable,hu2020what}.

\subsection{Why a Survey on Natural Language Generation?}
Here, we will explain the reasons and motivations why the natural language generation is worth reviewing and investigating.

Reiter \textit{et al.} \cite{reiter1997building} provided the most classical survey of NLG. However, the field of NLG has changed drastically in the last 20 years, with the emergence of successful deep learning methods. For example, since 2014, various neural encoder-decoder models pioneered by sequence-to-sequence (Seq2Seq) have been proposed to achieve the goal by learning to map input text to output text. In addition, the evaluation of NLG output should start to receive systematic attention.

Since Reiter \textit{et al.} \cite{reiter1997building} published their book, various other NLG overview texts have also appeared.
Gatter \textit{et al.} \cite{gatt2018survey} introduce the core tasks, applications and evaluation metrics of natural language generation. While useful, this survey is not highly timely and does not include the state-of-the-art research on the novel deep learning models such as graph neural networks.
Perera \textit{et al.} \cite{perera2017recent} cover some tasks or architectures of NLG.
Santhanam \textit{et al.} \cite{santhanam2019survey} review the NLG research progress of dialogue systems.
Mogandala \textit{et al.} \cite{mogadala2020trends} study the integration of vision and language in multimodal NLG, such as image dialogue and video storytelling.
Otter \textit{et al.} \cite{otter2020survey} concludes the progress of deep learning for NLP, display some classic text generation methods but barely discuss the research progress of NLG.
Yu \textit{et al.} \cite{yu2021survey} offers a survey of the knowledge-enhanced text generation methods.

The goal of the our survey is to present a highly timely overview of NLG developments from the aspect of data-to-text generation and text-to-text generation. Though NLG has been a part of AI and Natural language processing for long time, it has only recently begun to take full advantage of recent advances in data-driven, machine learning and deep learning approaches. Therefore this survey will focus on introducing the latest development and future directions of deep learning methods in field of NLG.
This survey can have broad audiences, researchers and practitioners, in academia and industry.

\subsection{Contribution of this Survey}
In this paper, we provide a thorough review of different natural language generation tasks as well as its corresponding datasets and methods. To summarize, this paper presents an extensive survey of natural language generations with the following contributions:
\begin{enumerate}
\item To give an up-to-date synthesis of deep learning research on the core tasks in NLG, as well as the architectures adopted in the field;
\item To detail meticulously and comprehensively various NLG tasks and datasets, and draw attention to the challenges in NLG evaluation, focusing on different evaluation methods and their relationships.
\item To highlight some future emphasis and relatively recent research issues that arise due to the increasing synergy between NLG and other artificial intelligence areas, such as computer vision, text and computational creativity.
\end{enumerate}

The rest of this survey is organized as follows. In Sec.\ref{General_Methods}, we introduce the general methods of NLG to give a comprehensive understanding. From Sec.\ref{text_abbreviation} to Sec.\ref{from_image_to_text_generation}, we will give a comprehensive introduction to the four main areas of NLG from the perspectives of task, data, and methods. In Sec.\ref{evaluation_metrics}, we present the important evaluation metrics used in various aforementioned NLG tasks. Besides, we propose some problems and challenges of NLG as well as several future research directions in Sec.\ref{problems_and_challenges}. And finally we conclude our survey in Sec.\ref{conclusions}.

\section{General Methods of NLG}
\label{General_Methods}

In general, the task of natural language generation (NLG) targets at finding an optimal sequence $y^{\*}_{<T+1} = (y_1, y_2, ..., y_T)$ that satisfies:
\begin{equation}
y^{\*}_{<T+1} = \argmax_{y_{<T+1} \in \mathcal{Y}} \log P_\theta (y_{<T+1} | x) = \argmax_{y_{<T+1} \in \mathcal{Y}} \sum_{t=1}^T \log P_\theta (y_t | y_{<t}, x),
\end{equation}
where $T$ represents the number of tokens of the generated sequence, $\mathcal{Y}$ represents a set containing all possible sequences, and $P_\theta (y_t | y_{<t},  x)$ is the conditional probability of the next token $y_t$ based on its previous tokens $y_{<t} = (y_1, y_2, ..., y_{t-1})$ and the source sequence $x$ with model parameters $\theta$.

The general methods to deal with the tasks of NLG mainly contain: Recurrent Neural Network, Transformer, Attention Mechanism, Copy and Pointing Mechanisms, Generative Adversarial Network, Memory Network, Graph Neural Network, and Pre-trained Model. 

\subsection{Recurrent Neural Network}
As proposed by \cite{sutskever2014sequence}, the encoder of the sequence-to-sequence (Seq2Seq) framework is an Recurrent Neural Network (RNN), it will traverse every token (word) of the input, the input of each time is the hidden state and input of the previous time, and then there will be an output and a new hidden state. The new hidden state will be used as the input hidden state of the next time. We usually only keep the hidden state of the last time, which encodes the semantics of the whole sentence. After the encoder processing, the last hidden state will be regarded as the initial hidden state of the decoder. The Decoder is also an RNN, which outputs one word at a time. The input of each time is the hidden state of the previous time and the output of the previous time. The initial hidden state is the last hidden state of the encoder, and the input is special. Then, RNN is used to calculate the new hidden state and output the first word, and then the new hidden state and the first word are used to calculate the second word. Until EOS is encountered and the output is finished. A standard RNN computes a sequence of outputs ($y_1, ..., y_T$) given a sequence of inputs ($x_1, ..., x_T$) by iterating the following equation:
\begin{equation}
y_t = W^{\hbox{yh}} \cdot h_t = W^{\hbox{yh}} \cdot \sigma(W^{\hbox{hx}} x_t + W^{\hbox{hh}} h_{t - 1}),
\end{equation}
where $\sigma$ is activation function, $W^{\hbox{hx}}, W^{\hbox{hh}}, W^{\hbox{yh}}$ are learnable parameters, and $h_t$ is the hidden state at $t$-th timestep.

\subsection{Transformer}
Transformer \cite{vaswani2017attention} is based on the encoder-decoder framework, and both encoder and decoder are composed of stacked identified layers. The encoder is used to map an input sequence of symbol representations to another sequence of continuous representations, and then the decoder auto-regressively generates an output sequence based on its previously generated symbols and the continuous representations from encoder. In the encoder, each layer contains two sub-layers, which are multi-head self-attention mechanism (MultiHeadAttn) and position-wise fully connected feed-forward network (FFN) respectively. The multi-head self-attention can be formulated by:
\begin{gather}
\hbox{MultiHeadAttn}(Q,K,V) = \hbox{Concat}(\hbox{head}_1, ..., \hbox{head}_h) W^O, \\
\hbox{head}_i = \hbox{Attention}(QW_i^Q, KW_i^K, VW_i^V), \\
\hbox{Attention}(Q,K,V) = \hbox{softmax}(\frac{QK^{\top}}{\sqrt{d_k}}) V,
\end{gather}
where $Q,K,V$ are the query, key, and value matrices, $d_k$ is the dimension of queries and keys. And the feed-forward network can be formulated by:
\begin{gather}
\hbox{FFN}(x) = \hbox{max}(0, xW_1 + b_1) W_2 + b_2.
\end{gather}
Each of the stacked layer is surrounded by a residual connection followed by layer normalization.  While in the decoder, each layer contains three sub-layers with an additional multi-head attention over the encoder's output. The self-attention in decoder is masked to prevent attending to subsequent tokens. In addition, to inject the position information, the position encodings are added to the input embeddings as formulated by:
\begin{equation}
PE_{pos, 2i} = sin(pos / 10000^{2i / d_{model}}), \ \ PE_{pos, 2i+1} = cos(pos / 10000^{2i / d_{model}}),
\end{equation}
where $pos$ is the position and $i$ is the dimension.

The most commonly used loss function is the conditional language modeling loss, which can be formulated as: 
\begin{equation}
L = \sum_{t=1}^T \log P_{\theta}(y_t|y_{<t}, x),
\end{equation}
where $\log P_{\theta}(y_t|y_{<t}, x)$ is the log-likelihood of the $t$-th generated token conditioned on the previously generated sequence $y_{<t}$ and the source sequence $x$.

\subsection{Attention Mechanism}
Attention mechanism \cite{chorowski2015attention-based} is used to perform a mapping from a query to a series of key value pairs. There are three steps in the calculation of attention. The first step is to calculate the similarity between query and each key to get the weight. The common similarity functions are dot product, splicing, perceptron, etc.; the second step is to normalize these weights by using a softmax function; the last step is to sum the weight and the corresponding key value to get the final attention. The tokens in an article are first fed into the encoder to generate a sequence of encoder hidden states $h_i$, then the decoder receives the word embedding of previous step and derive the decoder state $s_t$ at each step $t$. After that, the attentive context vector $h_t^{*}$ can be obtained based on the attention distribution $a^t$ as:
\begin{equation}
h_t^{*} = \sum_i a_i^t h_i, \text{ where } a_i^t = \hbox{softmax}(e_i^t), \ e_i^t = v^{\top} \hbox{tanh}(W_h h_i + W_s s_t + b_{attn}),
\end{equation}
in which $v, W_h, W_s, b_{attn}$ are learnable parameters. Finally, the predicted vocabulary distribution $P_{vocab}$ is obtained by:
\begin{gather}
P_{vocab} = \hbox{softmax}(V^{\prime} (V[s_t, h_t^{*}] + b) + b^{\prime}),
\end{gather}
where $V, V^{\prime}, b, b^{\prime}$ are learnable parameters.

\subsection{Copy and Pointing Mechanisms}
The copy and pointing mechanisms proposed by PGN \cite{see2017get} is widely used in abstractive summarization, which is designed for alleviating the problem of inaccurate reproduced factual details via a pointer, dealing with out-of-vocabulary words and repetition via a generator and a coverage mechanism, respectively. In the pointer and generator, a generation probability $p_{gen}$ based on the context vector $h_t^{*}$, decoder state $s_t$ and decoder input $x_t$ at step $t$ is calculated by:
\begin{gather}
p_{gen} = \sigma(w_{h^{*}}^{\top} h_t^{*} + w_s^{\top} s_t + w_x^{\top} x_t + b_{ptr}),
\end{gather}
which serves as a soft switch to choose between generating a word based on the vocabulary probability, or copying a word from the source document based on the attention distribution. The probability distribution over the extended vocabulary $P(w)$ can be formulated as:
\begin{gather}
P(w) = p_{gen} P_{vocab}(w) + (1-p_{gen}) \sum_{i:w_i=w} a_i^t.
\end{gather}
And in the coverage mechanism, a coverage vector $c^t$ at step $t$ is calculated by the sum of attention distributions over previous decoder timesteps as $c^t = \sum_{t^{\prime}=0}^{t-1} a^{t^{\prime}}$, which is then added to the attention mechanism and the primary loss function as:
\begin{align}
e_i^t &= v^{\top} \hbox{tanh}(W_h h_i + W_s s_t + w_c c_i^t + b_{attn}), \\
loss_t &= -\hbox{log} P(w_t^{*}) + \lambda \sum_i \min(a_i^t, c_i^t),
\end{align}
where $w_t^{*}$ is the target word at step $t$.

\subsection{Generative Adversarial Network}
The generative adversarial network (GAN) \cite{goodfellow2014generative} is a framework that uses an adversarial training process to estimate generative models. This framework can be regarded as a minimax two-player game containing a generative model and a discriminative model, and these two models are simultaneously trained. The generator (G) captures the data distribution and tries to produce fake samples, and the discriminator (D) attempts to determine whether the samples come from the model distribution or data distribution. In detail, G is trained to maximize the probability identified by D for the sample coming from the data rather than G, while D is trained to maximize the probability of assigning the correct label to training samples and samples generated by G. The training process continues until the counterfeits are indistiguishable from the genuine articles. The training objective with value function $V(D,G)$ can be formulated as:
\begin{equation}
\min\limits_{G} \max\limits_{D} V(D,G) = \mathbb{E}_{\mathbf{x} \sim p_{data}(\mathbf{x})} [\log D(\mathbf{x})] + \mathbb{E}_{\mathbf{z} \sim p_{\mathbf{z}}(\mathbf{z})} [\log (1 - D(G(\mathbf{z})))].
\end{equation}

\subsection{Memory Network}
The end-to-end memory network \cite{sukhbaatar2015end-to-end} is based on the recurrent neural network. Before outputting a symbol, the recurrence reads from a possibly large external memory multiple times. In this framework, a discrete set of input sentences is written to the memory up to a fixed size, and a continuous representation for each memory sentence and the query (i.e., the question) is calculated and is then processed through multiple hops to output the answer. Specifically, for the single hop operation, the matching probability between each query and memory is first computed by the inner product followed by a softmax in the embedding space. The input set $x$ and the query $q$ are first embedded to obtain the memory vectors $m$ and $u$, respectively, and the match probability $p_i$ of the $i$-th input sentence is calculated as: 
\begin{equation}
p_i = \hbox{softmax}(u^{\top} m_i).
\end{equation}
Then the output memory representation $o$ is obtained by a weighted sum over another embedded memory sentences $c$ based on the matching probability as:
\begin{equation}
o = \sum_i p_i c_i.
\end{equation}
Finally, the final prediction $\hat{a}$ can be obtained through a weight matrix and a softmax over the sum of output memory vector $o$ and input embedding $u$ as:
\begin{equation}
\hat{a} = \hbox{softmax}(W(o+u)).
\end{equation}
To handle multiple hop operations, the memory layers are repeatedly stacked, and the input of each layer is the sum of the output memory vector and the input from its previous layer.

\subsection{Graph Neural Network}
The graph neural network (GNN) \cite{scarselli2009the-graph} is used to process the data in graph form, which can capture the dependency information between nodes of a graph via message passing. Compared with CNN and RNN, GNN can propagate on each node respectively and is able to ignore the input orders of nodes, which is more computational efficient. The representation of each node in a graph is iteratively updated by aggregating information from its neighboring nodes and edges. So far, many propagation strategies have been proposed, such as convolution \cite{kipf2017semi-supervised}, RNN based gate mechanism \cite{li2016gated}, and attention mechanism \cite{velickovic2018graph}. Meanwhile, many works attempt to improve the training method, such as sampling-based training \cite{hamilton2017inductive} and unsupervised training \cite{kipf2016variational}. Take the graph convolution network \cite{kipf2017semi-supervised} as an example, it follows the layer-wise propogation rule as:
\begin{equation}
H^{(l+1)} = \sigma \left( \tilde{D}^{-\frac{1}{2}} \tilde{A} \tilde{D}^{-\frac{1}{2}} H^{(l)} W^{(l)} \right),
\end{equation}
where $\sigma (\cdot)$ refers to an activation fuction, $\tilde{A} = A + I_N$ is the adjacency matrix of the graph (including the identity matrix $I_N$), $\tilde{D}_{ii} = \sum_j \tilde{A}_{ij}$ is the degree of node $i$, $W^{(l)}$ is a trainable weight matrix in the $l$-th layer, and $H^{(l)}$ is the matrix of node feature vectors in the $l$-th layer.

\subsection{Pre-trained Model}
The pre-trained models can be divided into two categories: non-contextual and contextual. The non-contextual pre-trained models (e.g., Word2vec \cite{mikolov2013distributed}, GloVe \cite{pennington2014glove}) can learn high quality of word and phrase representations, and are widely used to initialize the embeddings and improve the model performance for generation tasks. However, since the non-contextual embeddings are static, it is unable to handle context-dependent words and out-of-vocabulary words. 

To overcome these problems, contextual pre-trained models are proposed, which can dynamically change the word embeddings when the word appears in different sentences. Traditional contextual models mainly focus on the tasks of natural language understanding (NLU), which are based on the architectures of LSTM (e.g., ELMo \cite{peters2018deep}) or Transformer encoder (e.g., BERT \cite{devlin2019bert}). 

To tackle the more challenging tasks of natural language generation (NLG), the architecture of Transformer decoder is widely adopted, and the pre-training objective is further modified to adapt to the NLG task. For example, GPT \cite{brown2020language} uses the standard language modeling as the pre-training objective. T5 \cite{raffel2020exploring} designs a unified text-to-text framework for both NLU and NLG. During pre-training, each corrupted token span in the input sequence is replaced with a sentinel token, and the output sequence is composed of the dropped-out spans. MASS \cite{song2019mass} predicts the sentence fragment with input of masked sequence during pre-training. UniLM \cite{dong2019unified} propose a unified pre-training framework for both NLU and NLG tasks, which is based on a shared Transformer model with three different types of self-attention masks to switch among different tasks, including unidirectional, bidirectional, and sequence-to-sequence. BART \cite{lewis2020bart} designs a reconstruction objective to restore corrupted documents, where five types of transformation strategies are proposed, including token masking, token deletion, text infilling, sentence permutation, and document rotation. PLATO \cite{bao2020plato} propose a pre-training framework targeting the dialogue generation tasks with two reciprocal pre-training tasks, i.e., response generation and latent act recognition. The latent discrete variables are also introduced to solve the one-to-many mapping problem. ERNIE-GEN \cite{xiao2020ernie-gen} is a multi-flow Seq2Seq pre-training model, which pays attention to the exposure bias problem of pre-training models on downstream NLG tasks such as question generation, and aims to make the NLG models generate more human-like and fluent texts. OFA \cite{wang2022unifying} present a unified multimodal pre-training framework to unify various vision and language tasks, such as NLU, NLG, and image classification. Transformer is applied as the backbone architecture, and the sparse coding and a unified vocabulary are utilized to represent the images and linguistic words.

\section{Text Abbreviation}\label{text_abbreviation}
\subsection{Task}
The goal of text abbreviation is to distill key information from long texts to short ones, which consists of three subtopics: text summarization, question generation, and distractor generation. 

There are two kinds of methods for text abbreviation: extractive and abstractive methods. Since the abstractive approach is more flexible and can create more human-like sentences than the extractive approach, it has been paid more and more attentions in recent years and is our main focus in this paper. Text summarization is the process of generating entirely new phrases and sentences to capture the meaning of the source document. Question generation concentrates on automatically generating questions from a given sentence or paragraph. Distractor generation is the automatic generation of adequate distractors for a given question answer pair generated from a given article to form an adequate multiple-choice question. We summarize the most representative methods for each subtask in Table~\ref{tab: text abbreviation}.

\begin{table*}
\centering
\caption{Natrual language generation models for text abbreviation.}\label{tab: text abbreviation}
\resizebox{\textwidth}{!}{
\begin{tabular}{l|l|l}
\toprule
\textbf{Task}& \textbf{Model}& \textbf{Description} \\

\midrule

\multirow{8}{*}{Text Summarization}& MASS \cite{song2019mass}& Transformer \\
~& BART \cite{lewis2020bart}& Transformer + Multi-task Learning \\
~& PEGASUS \cite{zhang2020pegasus}& Transformer + Multi-task Learning \\
~& RCT \cite{cai2019improving}& Transformer + RNN + CNN + Long-term Dependency \\
~& ProphetNet \cite{qi2020prophetnet}& Transformer + Long-term Dependency \\
~& En-Semantic-Model \cite{ding2020generative}& RNN + Long-term Dependency \\
~& Post-Editing Factual Error Corrector \cite{cao2020factual}& Transformer + Factual Consistency \\
~& SpanFact \cite{dong2020multi}& Transformer + Factual Consistency \\

\midrule

\multirow{13}{*}{Question Generation}& Key-Phrase-based Question Generator \cite{dw2021question}&  Keyphrase + Template \\
~& Dynamic Mathematical Question Generator \cite{bhatia2019dynamic}& Constraint Handling Rules \\
~& KB-based Factoid Question Generator \cite{serban2016generating}& RNN \\
~& Teacher Forcing and RL Based Question Generator \cite{yuan2017machine}& RNN + RL \\
~& Paragraph-level Question Generator \cite{zhao2018paragraph-level}& RNN \\
~& Answer-Position-aware Question Generator \cite{sun2018answer}& RNN + Answer-focused \\
~& ASs2s \cite{kim2019improving}& RNN + Answer-focused \\
~& NQG-MP \cite{wang2019a-multi-agent}& RNN + Multi-task Learning \\
~& Paraphrase Enhanced Question Generator \cite{jia2020how}& RNN + Multi-task Learning \\
~& CGC-QG \cite{liu2019learning}& RNN + Multi-task Learning + GNN \\
~& PathQG \cite{wang2020pathqg}& RNN + Multi-task Learning + KG \\
~& UniLM \cite{dong2019unified}& Transformer + Multi-task Learning \\
~& ERNIE-GEN \cite{xiao2020ernie-gen}& Transformer + Multi-task Learning \\
\midrule

\multirow{9}{*}{Distractor Generation}& Educational Ontology Distractor Generator \cite{stasaski2017multiple}& Ontology + Embedding \\
~& Learning to Rank Based Distractor Generator \cite{liang2018distractor}& Embedding + Ranking + GAN + RL \\
~& BERT-based Distractor Generation \cite{chung2020a-bert-based}& BERT + Multi-task Learning \\
~& Hierarchical Dual-attention Distractor Generator \cite{gao2019generating}& RNN \\
~& EDGE \cite{qiu2020automatic}& RNN + Answer Interaction \\
~& HMD-Net \cite{maurya2020learning}& Transformer + RNN + Answer Interaction \\
~& Code Compression Distractor Generator \cite{srinivas2019towards}& Abstract Syntax Tree \\
~& CSG-DS \cite{ren2021knowledge-driven}& LDA + KB + Ranking \\
~& Named Entity Distractor Generator \cite{patra2019a-hybrid}& Tree + Clustering \\
\bottomrule
\end{tabular}}
\end{table*}

\subsection{Data}
\subsubsection{Text Summarization}
There are mainly four datasets in the field of text summarization as shown below.

\paragraph{CNN/DailyMail} 
The CNN/DailyMail dataset \cite{hermann2015teaching} is a large scale reading comprehension dataset. This dataset contains 93k and 220k articles collected from the CNN and Daily Mail websites, respectively, where each article has its matching abstractive summary. 

\paragraph{NYT} The New York Times (NYT) dataset \cite{sandhaus2008the-new, paulus2018a-deep} contains large amount of articles written and published by the New York Times between 1987 and 2007. In this dataset, most of the articles are manually summarized and tagged by a staff of library scientists, and there are over 650,000 article-summary pairs. 

\paragraph{XSum} The extreme summarization (XSum) dataset \cite{narayan2018dont} is an extreme summarization dataset containing BBC articles and corresponding single sentence summaries. In this dataset, 226,711 Wayback archived BBC articles are collected, which range from 2010 to 2017 and cover a wide variety of domains. 

\paragraph{Gigaword} The English Gigaword dataset \cite{graff2003english, rush2015a-neural} is a comprehensive collection of English newswire text data acquired by the Linguistic Data Consortium. This corpus contains four distinct international sources of English newswire, and has totally 4,111,240 documents.

\subsubsection{Question Generation}
The two popular datasets for the task of question generation are shown below.

\paragraph{SQuAD} The Stanford Question Answering Dataset (SQuAD) \cite{rajpurkar2016squad} is a large reading comprehension dataset created by crowdworkers. The questions in this dataset are posed by crowdworkers based on a set of Wikipedia articles, and the answers are text segments from the corresponding passages. In total, SQuAD contains 107,785 question-answer pairs on 536 articles. 

\paragraph{MS MARCO} The Microsoft Machine Reading Comprehension (MS MARCO) dataset \cite{nguyen2016ms-marco} is a collection of anonymized search queries issued through Bing or Cortana for reading comprehension. The dataset contains both answerable and unanswerable questions. Each answerable question has a set of extracted passages from the retrieved response documents of Bing. There are totally 1,010,916 questions and 8,841,823 answering passages extracted from 3,563,535 web documents in this dataset.

\subsubsection{Distractor Generation}
There are three datasets widely used for distractor generation as shown below. 

\paragraph{SciQ} SciQ \cite{welbl2017crowdsourcing} is a crowdsourced multiple choice question answering dataset, which consists of 13.7k science exam questions. The domain of this dataset covers biology, chemistry, earth science, and physics.

\paragraph{MCQL} The MCQL dataset \cite{liang2018distractor} is a collection of multiple choice questions at the Cambridge O level and college level, which is crawled from the Web. This dataset totally contains 7.1k questions covering biology, phisics, and chemistry.

\paragraph{RACE} RACE \cite{lai2017race} is a reading comprehension dataset collected from the English exams in Chinese middle and high schools. This dataset contains 27,933 passages and 97,687 questions, covering all types of human articles.

\subsection{Method}
\subsubsection{Text Summarization}
Recently, the most common methods in this field are encoder-decoder based pre-trained language models. Song \textit{et al.} \cite{song2019mass} design a novel pre-training objective to jointly pre-train the encoder and decoder, where the decoder learns to predict the masked sentence fragments in the encoder side. Given an unpaired source sentence $x$ from the source domain $\mathcal{X}$, the model with parameter $\theta$ predicts the sentence fragment $x^{u:v}$ from position $u$ to $v$ with the masked sequence $x^{\setminus u:v}$ as input, and the objective function is formulated as:
\begin{equation}
L(\theta;\mathcal{X}) = \frac{1}{|\mathcal{X}|} \sum_{x \in \mathcal{X}} \log P(x^{u:v}|x^{\setminus u:v};\theta) = \frac{1}{|\mathcal{X}|} \sum_{x in \mathcal{X}} \log \prod \limits_{t=u}^v P(x_t^{u:v}|x_{<t}^{u:v}, x^{\setminus u:v};\theta).
\end{equation}
Lewis \textit{et al.} \cite{lewis2020bart} present a denoising autoencoder for pre-training, which consists of a series of noising strategies to corrupt text and a training objective to reconstruct the original sentence. In addition, Zhang \textit{et al.} \cite{zhang2020pegasus} propose a self-supervised pre-training objective specific for the text summarization task, namely the gap-sentences generation objective, and their PEGASUS model achieves state-of-the-art performances on the mainstream datasets.

However, the pre-trained language models' ability of capturing long-term dependencies and maintaining global coherence is poor. To solve this problem, Cai \textit{et al.} \cite{cai2019improving} introduce an additional encoder with a bidirectional RNN and a convolution module to simultaneously model sequential context and capture local importance to the base Transformer model. The encoder first applies a bidirectional LSTM on the source text embedding $E$ and derive the hidden states sequence $H$, then uses three convolution operations with kernel sizes of 1,3,5 to learn n-gram features $D$ from $H$, and finally a gated linear unit is applied to select features, which can be formulated by:
\begin{equation}
R = \sigma (W_d D + b_d) \odot (W_h H + b_h),
\end{equation}
where $H = \hbox{BiLSTM}(E)$. Yan \textit{et al.} \cite{qi2020prophetnet} make several improvements to the traditional language models. Specifically, a new self-supervised objective called future n-gram prediction has been applied, and the n-stream self-attention mechanism is used in the decoder to be trained to predict the future n-gram of each time step. Given a source sequence $x$ and its target sequence $y$, the future n-gram prediction objective can be formulated as:
\begin{equation}
\mathcal{L} = -\sum_{j=0}^{n-1} \alpha_j \cdot \left( \sum_{t=1}^{T-j} \log p_{\theta} (y_{t+j}|y_{<t}, x) \right).
\end{equation}
Ding \textit{et al.} \cite{ding2020generative} propose an enhanced semantic model based on double encoders and a decoder with a Gain-Benefit gate structure, which is able to provide richer semantic information and reduce the influence of the length of generated texts on the decoding accuracy. The dual-encoder is used to capture both global and local context semantic information based on the bidirectional RNN, and the output vector of the Gain-Benefit gate module is calculated by:
\begin{equation}
\mathbf{P}_t = (1 - a) \odot \mathbf{M} + a \odot \mathbf{C}_{t - 1}, \text{ where } a = \hbox{sigmoid} (\mathbf{W}_1 \mathbf{C}_t + \mathbf{W}_2 \mathbf{S}_{t - 1} + y_{t - 1}),
\end{equation}
in which $\mathbf{C}_t$ is the contextual semantic representation with both global and local semantic information at current time step, $\mathbf{S}_{t-1}$ is the decoder hidden state of last time step, $y_{t - 1}$ is the word generated at last time step, and $\mathbf{M}$ is the global semantic vector of the original text.

Except for the poor capability of capturing long-term dependency in traditional pre-trained language models, the problem of factual inconsistency between the generated content and source text is also severe and has not been tackled by previous works. To reduce this phenomenon, Meng \textit{et al.} \cite{cao2020factual} propose an end-to-end neural corrector model with a post-editing correction strategy, which is pre-trained on artificial corrupted reference summaries. Dong \textit{et al.} \cite{dong2020multi} introduce a factual correction framework containing a QA-span factual correction model and an auto-regressive one. The QA-span correction model masks and replaces one entity at a time during the iteration process. Specifically, given the source text $x$ and a masked query $q=(y_1^{\prime}, ..., [MASK], ..., y_m^{\prime})$, the correction model needs to predict the answer span via $p(i=start)$ and $p(i=end)$, which are calculated based on the hidden states of the top layer $h_i$ as:
\begin{equation}
p(i=start) = a_i^{start} = \frac{\hbox{exp}(q_i^s)}{\sum_{j=0}^{H-1} \hbox{exp}(q_j^s)}, \text{ where } q_i^s = \hbox{ReLU}(w_s^{\top} h_i + b_s),
\end{equation}
in which $H$ is the number of hidden states in the encoder, and $p(i=end)$ is calculated in the similar way. The auto-regressive correction model masks all entities at the same time without iteration. Specifically, given the source text $x$ and a masked query $q=(y_1^{\prime}, ..., [MASK]_1, ..., [MASK]_T, ..., y_m^{\prime})$ from a summary with $T$ entities, the correction model runs $T$ steps to predict the answer span of each mask based on the corresponding masked token representation and its previously predicted entity representations. The entity representation $\mathbf{s}_t^{ent}$ at time step $t$ is predicted based on the argmax and mean pooling operations, as calculated by:
\begin{equation}
\mathbf{s}_t^{ent} = \hbox{Mean-Pool}({\mathbf{h}_{p_{start}}, \mathbf{h}_{p_{end}}}),
\end{equation}
where $p_{start} = \argmax(a_1^{start}, ..., a_M^{start}), p_{end} = \argmax(a_1^{end}, ..., a_M^{end})$.

\subsubsection{Question Generation}
Early methods to solve the task of question generation are mainly based on hand-crafted rules and always contain multiple procedures. Wijanarko \textit{et al.} \cite{dw2021question} propose a method that generates questions based on key-phrase which embeds Bloom’s taxonomic in selecting contexts for constructing questions. The key-phrases can be selected via two methods. The first method is based on probability of two-word sequence as formulated below:
\begin{equation}
score(w_i, w_j) = \frac{count(w_i w_j) - \delta}{count(w_i) \times count(w_j)},
\end{equation}
where $count(w_i, w_j)$ is the number of a sequence of word $w_i$ followed by $w_j$, $count(w_i)$ is the number of word $w_i$ in the input document, and $\delta$ is a constant to limit the number of phrases formed by less frequent words. The second method is based on a Naive Bayes model, and the probability of a unique phrase is measured by:
\begin{equation}
Pr[key|T, D] = \frac{Pr[T|key] \times Pr[D|key] \times Pr[key]}{Pr[T, D]},
\end{equation}
where $Pr[T|key]$ is the probability that a key-phrase has a $TF \times IDF$ score $T$, $Pr[D|key]$ is the probability that it has a distance $D$, and $Pr[T, D]$ is the normalization factor. Bhatia \textit{et al.} \cite{bhatia2019dynamic} propose a dynamic question answering generator to generate questions and answers for the mathematical topic-quadratic equations. The randomization technique, first order logic and automated deduction are used for the study. 

In the past few years, driven by advances in deep learning, end-to-end neural models based on Seq2Seq framework have attained more and more attentions and have shown better performances. Serban \textit{et al.} \cite{serban2016generating} study the factoid questions generation with RNN to transduce facts into neural language questions, and provide an enormous question-answer pair corpus. Yuan \textit{et al.} \cite{yuan2017machine} use a Seq2Seq model with teacher forcing to improve the training and adopts policy gradient in reinforcement learning to optimize the generated results. During supervised learning, in addition to minimize the negative log-likelihood with teacher forcing, two additional signals are introduced to prevent the model from generating answer words ($\mathcal{L}_s$) and encourage the output variety ($\mathcal{L}_e$) as formulated below:
\begin{equation}
\mathcal{L}_s = \lambda_s \sum_t \sum_{\bar{a} \in \bar{\mathcal{A}}} p_{\theta} (y_t = \bar{a} | y_{<t}, D, A), \ \ \mathcal{L}_e = \lambda_e \sum_t \mathbf{p}_t^T \hbox{log} \mathbf{p}_t, 
\end{equation}
where $\bar{\mathcal{A}}$ refers to the set of words appearing in the answer but not in the ground-truth question, $\mathbf{p}_t$ refers to the full pointer-softmax probability of the $t$-th word, which enables the model to interpolate between copying from the source document and generating from shortlist. And during reinforcement learning, the total reward $R_{PPL+QA}$ is a combination of question answering reward $R_{QA}$ (measured by F1 score) and question fluency reward $R_{PPL}$ (measured by perplexity), as formulated below:
\begin{align}
R_{PPL+QA} &= \lambda_{QA} R_{QA} (\hat{Y}) + \lambda_{PPL} R_{PPL} (\hat{Y}), \\
R_{PPL}(\hat{Y}) &= -2^{-\frac{1}{T} \sum_{t=1}^{T} \log_2 PLM(\hat{y}_t|\hat{y}_{<t})}, \\
R_{QA}(\hat{Y}) &= \hbox{F1}(\hat{A}, A),
\end{align}
where $\hat{A} = MPCM(\hat{Y})$ refers to the answer of the generated question by the Multi-Perspective Context Matching (MPCM) model, $PLM$ is a language model. Zhao \textit{et al.} \cite{zhao2018paragraph-level} study the paragraph-level neural question generation by proposing a maxout pointer and gated self-attention networks, which mainly deals with the problem that long text (mostly paragraphs) does not perform well in the Seq2Seq model. In detail, the gated self-attention network contains two steps: 1) the encoded passage-answer representation $\mathbf{u}_t$ is conducted matching against itself to derive the self matching representation $\mathbf{s}_t$ at time step $t$:
\begin{align}
\mathbf{s}_t &= \mathbf{U} \cdot \mathbf{a}_t^s = \mathbf{U} \cdot \hbox{softmax}(\mathbf{U}^{\top} \mathbf{W}^s \mathbf{u}_t).
\end{align}
And 2) the self matching representation $\mathbf{s}_t$ is combined with the original passage-answer representation $\mathbf{u}_t$, which is then fed into a feature fusion gate to obtain the final encoded passage-answer representation $\mathbf{\hat{u}}_t$ at time step $t$:
\begin{equation}
\mathbf{\hat{u}} = \mathbf{g}_t \odot \mathbf{f}_t + (1 - \mathbf{g}_t) \odot \mathbf{u}_t, \text{ where } \mathbf{f}_t = \hbox{tanh}(\mathbf{W}^f [\mathbf{u}_t, \mathbf{s}_t]), \ \mathbf{g}_t = \hbox{sigmoid}(\mathbf{W}^g [\mathbf{u}_t, \mathbf{s}_t]).
\end{equation}

However, the rich information lying in the answer has not been fully explored, which leads to generating low-quality questions (e.g., having mismatched interrogative words with the answer type, copying context words far and irrelevant to the answer, including words from the target answer, and so on). To solve the problem, Sun \textit{et al.} \cite{sun2018answer} propose an answer-focused and position-aware model. It incorporates the answer embedding to explicitly generate a question word matching the answer type, and designs a position-aware attention mechanism by modeling the relative distance with the answer, which guides the model to copy the context words that are more close and relevant to the answer. Kim \textit{et al.} \cite{kim2019improving} propose to separate the target answer from the original passage, in order to avoid the generated question copying words from the answer. Specifically, it replaces the answer with a mask token, and introduces a keyword-net to extract key information from the answer. Given the encoded answer representation $h^a$ and the context vector $c_t$ of current decoding step, the keyword feature of each layer of the keyword-net $o_t^l$ can be formulated as:
\begin{equation}
o_t^l = \sum_j p_{tj}^l h_j^a, \text{ where } p_{tj}^l = \hbox{softmax}((o_t^{l-1})^{\top} h_j^a),
\end{equation}
in which $o_t^0$ is initialized by $c_t$. And the decoding hidden state $s_t$ of current timestep is calculated by:
\begin{equation}
s_t = \hbox{LSTM}(y_{t-1}, s_{t-1}, c_t, o_t^L),
\end{equation}
where $y_{t-1}$ is the output token of previous timestep, $L$ is the layer number of the keyword-net.

Moreover, many works recently attempt to conduct multi-task learning with external related tasks to further enhance the performance of question generation. 
Wang \textit{et al.} \cite{wang2019a-multi-agent} introduce a message passing mechanism to simultaneously learn the tasks of phrase extraction and question generation, which helps the model be aware of question-worthy phrases that are worthwhile to be asked about. 
Jia \textit{et al.} \cite{jia2020how} conduct multi-task learning with paraphrase generation and question generation, which can diversify the question patterns of the question generation module. During training, the weights of the encoder are shared by all tasks while those of the first layer of decoder are shared with a soft sharing strategy, which is formulated by:
\begin{equation}
\mathcal{L}_{sf} = \sum_{\mathbf{d} \in \mathcal{D}} \| \theta_d - \phi_d \|_2,
\end{equation}
where $\mathcal{D}$ is a set of shared decoder parameters, $\theta, \phi$ refer to the parameters of the question generation task and paraphrase generation task, respectively. And a min-loss function is employed among the golden reference question and several expanded question paraphrases, which is represented by:
\begin{equation}
\mathcal{L}_{qg} = \min_{\mathbf{q} \in \mathcal{Q}} (-\frac{1}{T_{qg}} \sum_{t=1}^{T_{qg}} log P(y_t^{qg} = \mathbf{q}_t)).
\end{equation}

Despite the outstanding performance achieved by previous methods, the rich structure information hidden in the passage is ignored, which can be used as an auxiliary knowledge of the unstructured input text to improve the performance. Liu \textit{et al.} \cite{liu2019learning} adopt a graph convolutional network (GCN) to identify the clue words in the input passage that should be copied into the target question. Specifically, the GCN is constructed on the syntactic dependency tree representation of each passage, and a Gumbel-Softmax layer is applied to the final representation of GCN to sample the binary clue indicator for each word. A sample $\mathbf{y} = (y_1, ..., y_k)$ drawn from the Gumbel-Softmax distribution is formulated as:
\begin{equation}
y_i = \frac{\exp ((\log (\pi_i) + g_i) / \tau)}{\sum_{j=1}^k \exp ((\log (\pi_j) + g_j) / \tau)},
\end{equation}
where $\tau$ is the temperature parameter, $\pi_i$ is the unnormalized log probability of class $i$, and $g_i$ is the Gumbel noise formulated by:
\begin{equation}
g_i = -\log (-\log (u_i)), \text{ where } u_i \sim \hbox{Uniform} (0,1).
\end{equation}
Wang \textit{et al.} \cite{wang2020pathqg} construct a knowledge graph for each input sentence as the auxiliary structured knowledge and aims to generate a question based on a query path from the knowledge graph. The query representation learning is formulated as a sequence labeling problem for identifying the involved facts to form a query, which is used to generate more relevant and informative questions. 

Recently, pre-trained language models have achieved remarkable performances in question generation, far exceeding those of the previous RNN-based methods. Dong \textit{et al.} \cite{dong2019unified} propose a unified model that is pre-trained with three types of natural language understanding or generation tasks, namely unidirectional, bidirectional, and sequence-to-sequence prediction. Through the pre-training of these three tasks, the model's question generation performance achieve significant improvements on SQuAD. Xiao \textit{et al.} \cite{xiao2020ernie-gen} propose an enhanced multi-flow seq2seq pre-training and fine-tuning framework to alleviate the exposure bias, which consists of an infilling generation mechanism and a noise-aware generation method, which achieves state-of-the-art performances on a wide range of datasets.

\subsubsection{Distractor Generation}
The researches mainly focus on generating multi-choice question distractors for ontologies or articles. Traditional methods primarily use hand-crafted rules or ranking method for distractor generation. Stasaski \textit{et al.} \cite{stasaski2017multiple} introduce a novel method with several ontology- and embedding-based approaches. The graph structure of the ontology is used to create complex problems linking different concepts. Liang \textit{et al.} \cite{liang2018distractor} introduce a ranking method with a feature-based model and a neural net (NN) based model. The NN-based model consists of a generator $G$ and a discriminator $D$, where $G$ generates distractors $d$ based on a conditional probability $P(d|q,a)$ given question stems $q$ and answers $a$, and $D$ predicts whether a distractor sample comes from the real training data or $G$. The objective for $D$ is to maximize the log-likelihood as:
\begin{equation}
\hbox{max}_{\phi} \mathbb{E}_{d \sim P_{true}(d|q,a)}[\log (\sigma (f_{\phi} (d|q,a)))] + \mathbb{E}_{d \sim P_{\theta} (d|q,a)} [\log (1 - \sigma (f_{\phi} (d|q,a)))],
\end{equation}
where $f_{\phi} (d,q,a)$ is an arbitrary scoring function parameterized by $\phi$. And each distractor $d_i$ sampled by $G$ is based on another scoring function $f_{\theta} (d,q,a)$ as formulated as:
\begin{equation}
p_{\theta}(d_i|q,a) = \frac{\hbox{exp}(\tau \cdot f_{\theta} (d_i, q, a))}{\sum_j \hbox{exp} (\tau \cdot f_{\theta} (d_j, q, a))},
\end{equation}
where $\tau$ is a temperature hyper-parameter. Then a cascaded learning framework is proposed to make the ranking more effective, which divides the ranking process into two stages to reduce the candidates. 

Recently, deep learning-based models are widely adopted due to its overwhelming performance. For example, Chung \textit{et al.} \cite{chung2020a-bert-based} utilize the BERT model to generate distractor with the auto-regressive mechanism in a multi-tasking architecture. Additionally, Gao \textit{et al.} \cite{gao2019generating} express the task as a sequence-to-sequence learning problem based on a hierarchical encoder-decoder network. In this model, static and dynamic attention mechanisms are adopted on the top of the hierarchical encoding structure, and a question-based initializer is used as the start point to generate distractors in the decoder. The question $q$, the answer $a$ and the word vectors in the $i$-th sentence $(\mathbf{w}_{i,1}, ..., \mathbf{w}_{i,m})$ are first encoded via three separate bidirectional LSTM networks into $(\mathbf{q}_1, ..., \mathbf{q}_l)$, $(\mathbf{a}_1, ..., \mathbf{a}_k)$, and $(\mathbf{h}_{i, 1}^e, ..., \mathbf{h}_{i, m})$. Then another bidirectional LSTM is applied on the encoded word representations to derive the contextualized sentence representation $(\mathbf{u}_1, ..., \mathbf{u}_n)$, and an average pooling layer is applied to derive their entire representations $\mathbf{q}$, $\mathbf{a}$, and $\mathbf{s}_i$. After that, the static attention distribution $\gamma_i$ can be derived through a matching layer and a normalization layer as:
\begin{equation}
\gamma_i = \hbox{softmax}(o_i/\tau), \text{ where } \tau = \hbox{sigmoid}(\mathbf{w_q}^{\top} \mathbf{q} + b_q), \ o_i = \lambda_q \mathbf{s}_i^{\top} \mathbf{W}_m \mathbf{q} - \lambda_a \mathbf{s}_i^{\top} \mathbf{W}_m \mathbf{a} + \mathbf{b}_m.
\end{equation}
In the decoder size, the decoder generates the hidden state $\mathbf{h}_t^d$ at the $t$-th time step through an LSTM network. Then the sentence-level and word-level dynamic attention $\beta_i$ and $\alpha_{i,j}$ are formulated as:
\begin{equation}
\beta_i = \mathbf{u}_i^{\top} \mathbf{W}_{d_1} \mathbf{h}_t^d, \ \  \alpha_{i,j} = {\mathbf{h}_{i,j}^e}^{\top} \mathbf{W}_{d_2} \mathbf{h}_t^d.
\end{equation}
Finally, the static and dynamic attentions are combined into $\tilde{\alpha}_{i,j}$ to reweight the article token representations and predict the probability distribution $P_V$ over vocabulary $V$:
\begin{equation}
P_V = \hbox{softmax}(\mathbf{W}_V \tilde{\mathbf{h}}_t^d + \mathbf{b}_V), \text{ where } \tilde{\mathbf{h}}_t^d = \hbox{tanh}(\mathbf{W}_{\tilde{\mathbf{h}}} [\mathbf{h}_t^d; \mathbf{c}_t]), \ \mathbf{c}_t = \sum_{i,j} \tilde{\alpha}_{i,j} \mathbf{h}_{i,j}^e, \ \tilde{\alpha}_{i,j} = \frac{\alpha_{i,j} \beta_i \gamma_i}{\sum_{i,j} \alpha_{i,j} \beta_i \gamma_i}.
\end{equation}

However, the answer interaction is not considered by previous works and the incorrectness of the generated distractors cannot be guaranteed. To address this problem, Qiu \textit{et al.} \cite{qiu2020automatic} propose a framework consisting of reforming modules and an attention-based distractor generator, which is the state-of-the-art method on most widely adopted datasets (e.g., RACE). The reforming modules use the semantic distances to constrain the effect of words that are strongly related to the correct answer, and the distractor generator leverages the information of the reformed question and passage to generate the initial state and context vector respectively. In detail, three contextual encoders are first applied to encode the passage, question and its answer into $\mathbf{P}$, $\mathbf{Q}$ and $\mathbf{A}$, then an attention mechanism and a fusion kernel are leveraged to enrich the question and answer representations into $\tilde{\mathbf{Q}}$ and $\tilde{\mathbf{A}}$, where $\tilde{\mathbf{Q}}$ is formulated by:
\begin{gather}
\tilde{\mathbf{Q}} = \hbox{Fuse}(\mathbf{Q}, \bar{\mathbf{Q}}) = \hbox{tanh}([\mathbf{Q}; \bar{\mathbf{Q}}; \mathbf{Q} - \bar{\mathbf{Q}}; \mathbf{Q} \circ \bar{\mathbf{Q}}] \mathbf{W}_f + \mathbf{b}_f), \\
\bar{\mathbf{Q}} = \hbox{Attn}(\mathbf{Q}, \mathbf{P}) \mathbf{P} = \hbox{softmax}(\frac{\mathbf{Q} \mathbf{P}^T}{\sqrt{d}}) \mathbf{P}.
\end{gather}
In the reforming question module, the reformed question $\dot{\mathbf{Q}}_i$ is calculated through a self-attend layer and a gate layer as:
\begin{align}
\dot{\mathbf{Q}} &= \hbox{Gate}(\tilde{\mathbf{Q}}_i, \tilde{\mathbf{v}}^a) \tilde{\mathbf{Q}}_i = (\tilde{\mathbf{Q}}_i \mathbf{W}_g^q \tilde{\mathbf{v}}^{a^{\top}} + b_g^q) \tilde{\mathbf{Q}}_i, \\
\tilde{\mathbf{v}}_a &= \hbox{SelfAlign}(\tilde{\mathbf{A}}) = \hbox{softmax}(\tilde{\mathbf{A}} \mathbf{W}_a)^{\top} \tilde{\mathbf{A}}.
\end{align}
And in the reforming passage module, the reformed passage $\tilde{\mathbf{P}}$ is calculated by:
\begin{gather}
\tilde{\mathbf{P}} = \hbox{Fuse}(\dot{\mathbf{P}}, \bar{\mathbf{P}}), \ \bar{\mathbf{P}} = \hbox{Attn}(\dot{\mathbf{P}}, \dot{\mathbf{Q}}) \dot{\mathbf{Q}}, \ \dot{\mathbf{P}}_i = \hbox{Gate}(\mathbf{P}_i, \hat{\mathbf{v}}^a) \mathbf{P}_i, \\
\hat{\mathbf{v}}^a = \hbox{SelfAlign}(\hat{\mathbf{A}}), \ \hat{\mathbf{A}} = \hbox{Fuse}(\tilde{\mathbf{A}}, \bar{\mathbf{A}}), \ \bar{\mathbf{A}} = \hbox{Attn}(\tilde{\mathbf{A}}, \tilde{\mathbf{Q}}) \tilde{\mathbf{Q}}.
\end{gather}

Maurya \textit{et al.} \cite{maurya2020learning} use a single encoder to encode the input triplet and three decoders to generate three distractors. The encoder employs SoftSel operation and a gated mechanism to capture the semantic relations among the elements of the input triplet. 

In addition, there are also many other scenarios for distractor generation. Srinivas \textit{et al.} \cite{srinivas2019towards} develop a semi-automatic tool to help teachers quickly create multiple choice questions on code understanding. The tool first captures each code structure in the form of an abstract syntac tree, and then train a code model that maps functions to vectors. Ren \textit{et al.} \cite{ren2021knowledge-driven} create a model based on a context-sensitive candidate set generator and a distractor selector for cloze-style multiple choice questions. The model first uses the correct answer as the key, combines the LDA model to mine the topic of the context, finds similar words in the semantic database, and then selects a specified number of misleading items according to a ranking model. Specifically, the probability distribution over all entites subsumed by the concepts in $C$ is calculated based on the posterior probability $p(c|a,q)$ as:
\begin{equation}
p_i = p(d_i|a,q) \propto \sum_{c \in C} p(d_i|c) p(c|a,q), \text{ where } p(c|a,q) \propto p(c|a) \sum_{k=1}^K \pi_{a,q}^{(k)} \gamma_c^{(k)},
\end{equation}
in which $c$ is the concept, $\pi_{a,q}$ is the topic distribution of complete sentence formed by the stem and key, $\gamma_c$ is the topic distribution of concept $c$, $p(c|a)$ is the prior probability of $a$ belonging to $c$, $K$ is the total number of topics, and $p(d|c)$ is the typicality. Patra \textit{et al.} \cite{patra2019a-hybrid} develop a system to generate named entity distractors. The system performs two types of similarity computation namely statistical and semantic. To speed up the distractor selection procedure, a hierarchical clustering method is proposed to represent the entities, where the entity similarities are embedded in a tree structure. The distractors are selected from the nearby entities of the correct answer of the question in the tree. Specifically, the statistical distance between the numeric attributes is calculated by:
\begin{equation}
Sim(P,Q) = 1 - \frac{1}{L} \sum_{i=1,...,L} \frac{(P_i \sim Q_i)}{\max (P_i, Q_i)},
\end{equation}
where $P$ and $Q$ represent two vectors corresponding to the target entities, $L$ is the total number of numeric attributes. The hierarchical distance between two entities ($x, x^{\prime}$) is normalized by:
\begin{equation}
Sim(x,x^{\prime}) = \frac{d_1(x,x^{\prime})}{\sqrt{d_1(x,x) \odot d_1(x^{\prime}, x^{\prime})}},
\end{equation}
where $d_1(x,x^{\prime})$ is the highest tree level connecting $x$ and $x^{\prime}$. And the semantic similarity score between the key $x$ and a candidate distractor $x^{\prime}$ is formulated as:
\begin{equation}
Sim(x,x^{\prime}) = \frac{|(triplet_i \in x)| \& (triplet_i \in x^{\prime})|_i}{|triplet_j \in x|_j},
\end{equation}
where the normalization factor is the size of the triplet set corresponding to the key.

\section{Text Expansion}\label{text_expansion}
\begin{table*}
\centering
\caption{Natrual language generation models for text expansion.}\label{tab: text expansion}
\resizebox{\textwidth}{!}{
\begin{tabular}{l|l|l}
\toprule
\textbf{Task}& \textbf{Model}& \textbf{Description} \\

\midrule

\multirow{4}{*}{Short Text Expansion}& Associated-Query-based Query Expander \cite{billerbeck2003query}& Ranking \\
~& ExpaNet \cite{tang2017end-to-end}& Retrieval + Memory Network \\
~& FC-LSTM \cite{shi2019functional}& LDA + RNN + Ranking \\
~& Fiction Sentence Expander \cite{safovish2020fiction}& RNN \\

\midrule

\multirow{5}{*}{Topic-to-Essay Generation}& MTA-LSTM \cite{feng2018topic-to-essay}& RNN + Global Coherence \\
~& SRENN \cite{wang2020self-attention}& RNN + Retrieval + Global Coherence \\
~& UD-GAN \cite{yuan2019efficient}& RNN + GAN + RL \\
~& KB-based Topic-to-essay Generator \cite{yang2019enhancing}& CNN + RNN + GAN + RL + KB \\
~& SCTKG \cite{qiao2020a-sentiment-controllable}& CNN + RNN + GAN + RL + KG \\
\bottomrule
\end{tabular}}
\end{table*}

\subsection{Task}
The main purpose of text expansion is to inflate the short texts to longer ones that contains more abundant information, which can be divided into two aspects: short text expansion and topic-to-essay generation. Short text expansion aims to expand a short text into a richer
representation based on a set of long documents. Topic-to-essay generation aims at generating human-like diverse, and topic-consistent paragraph-level text with a set of given topics. The most representative methods for each subtask are shown in Table~\ref{tab: text expansion}.

\subsection{Data}
\subsubsection{Short Text Expansion}
There are mainly four datasets for short text expansion as listed below.

\paragraph{Wikipedia} \cite{tang2017end-to-end} constructs this dataset through a snapshot of English Wikipedia. The titles of Wikipedia articles are regarded as short texts, and the abstract of all Wikipedia articles are leveraged to construct the related long documents. This dataset contains 30,000 short texts and 4,747,988 long documents in total. 

\paragraph{DBLP} \cite{tang2017end-to-end} uses the DBLP bibliography database to construct this dataset. The titles of computer science literature represent short texts, and the abstracts of all papers are collected to construct the corresponding long documents. Statistically, this dataset consists of 81,479 short texts and 480,558 long documents.

\paragraph{Programmableweb} \cite{shi2019functional} establishes a real-world dataset for service recommendation and description expansion, which is crawled from programmableweb.com in 2013 and 2016. The entire dataset contains 16012 APIs, 7816 Mashups and 16449 links between them. 

\paragraph{Fiction Corpus} \cite{safovish2020fiction} creates an English fiction corpus, which is obtained by applying sentence compression techniques on a modern fiction corpus scraped from online resources. Each story sentence has a corresponding compression. In total, the dataset contains around 17,000,000 sentences.

\subsubsection{Topic-to-Essay Generation}
There are primarily five datasets for topic-to-essay generation as listed below. 

\paragraph{ESSAY} The ESSAY dataset \cite{feng2018topic-to-essay} is a large collection of topic compressions crawled from the Internet. The topic words are extracted by TextRank. This datast totally contains 305,000 paragraph-level essays. 

\paragraph{ZhiHu} \cite{feng2018topic-to-essay} constructs this dataset by crawling from a Chinese question-and-anwering website called ZhiHu, which consists of 55,000 articles. The topic words of each article are specified by users in the community.

\paragraph{Moview Reviews} The Stanford Sentiment Treebank (SST) dataset \cite{socher2013recursive} contains 11,855 single sentences of movie reviews. This dataset has two sentiment classes for each review, and has a set of fully labeled parse trees. 

\paragraph{Beer Reviews} \cite{mcauley2013from} creates this dataset by crawling from the beer review website BeerAdvocate. In total, this dataset has 1,586,259 ratings on 66,051 items that are scored by 33,387 users. 

\paragraph{Customer Reviews} \cite{hu2004mining} constructs a customer review dataset consisting of five electronics products, which is collected from Amazon.com and C$\mid$net.com. There are totally 1,886 items in this dataset.

\subsection{Method}
\subsubsection{Short Text Expansion}
Early works primarily use statistical similarity for short text expansion. Billerbeck \textit{et al.} \cite{billerbeck2003query} propose a method about the query expansion using associated queries for web search engines. The main method is to associate a query that closely matches the document in a given log containing a large number of queries. Then, the query associated with the query document is a reasonable description of the document and can be used to expand the query text.

Benefit from the development of deep learning, many end-to-end frameworks based on neural networks have been developed. Tang \textit{et al.} \cite{tang2017end-to-end} propose an end-to-end solution based on deep memory network for short text extension. In this work, the original short text $q$ is firstly used as a query to search for a set of potentially relevant long documents $C_{q}$, which will be used as the material for text expansion, from an external large collection $C$. In the following process, the short text $q$ is represented as the average vector of words in it, i.e. $\overrightarrow{q}$, and each document is also represented by the average vector of words belong to it, i.e. $\overrightarrow{d}$. To further identify the relevant documents in $C_{q}$, both soft attention and hard attention mechanism are utilized, the information read from the document set can be written as:
\begin{equation}
\overrightarrow{o}=\sum_{i=1}^{K} p_{i} \cdot \overrightarrow{d_{i}} = \sum_{i=1}^{K} \mathrm{softmax}(\frac{\overrightarrow{q}^{T}\overrightarrow{d_{i}}+g_{i}}{\tau}) \cdot \overrightarrow{d_{i}},
\end{equation}
where $g_{i}$ follows the Gumbel(0,1) distribution, and $\tau$ is the temperature hyperparameter. Then the two sources of information, $\overrightarrow{q}$ and $\overrightarrow{o}$ are integrated by GRU as follows:
\begin{equation}
\overrightarrow{z}=\sigma (\textbf{W}^{(z)}\overrightarrow{q}+\textbf{U}^{(z)}\overrightarrow{o}),
\end{equation}
\begin{equation}
\overrightarrow{r}=\sigma (\textbf{W}^{(r)}\overrightarrow{q}+\textbf{U}^{(r)}\overrightarrow{o}),
\end{equation}
\begin{equation}
{\overrightarrow{o}}'=\textrm{tanh}(\textbf{W}\overrightarrow{q}+\overrightarrow{r}\circ \textrm{U}\overrightarrow{o}),
\end{equation}
\begin{equation}
{\overrightarrow{q}}'=(1-\overrightarrow{z})\circ \overrightarrow{q}+\overrightarrow{z}\circ {\overrightarrow{o}}',
\end{equation}
where $\circ$ means element-wise multiplication, both the Sigmoid function $\sigma(x)$ and $tanh(x)$ are operated on element-wise. The output ${\overrightarrow{q}}'$ is the expanded representation of the input short text $q$. This process can be repeated several times for the same $q$ to continuously extend its representation to mimic the human behavior when querying a piece of short text.

Shi \textit{et al.} \cite{shi2019functional} present a text expansion method for service recommendation system. The description of services is first expanded at sentence level by a probabilistic topic model, in this process, the similarity between target sentence $u$ and another sentence $v$ from existed corpus can be caculated as:
\begin{equation}
Similarity(u,v)=\mu \cdot D_{JS}(u,v)+(1-\mu)\cdot D_{JS}(S_{u},S_{v}),
\end{equation}
where $S_{u}$ and $S_{v}$ are descriptions contain $u$ and $v$, respectively. $\mu$ is a parameter used to balance weights of sentence and description on the final similarity measurement. $D_{JS}$ is the function of JS divergence which used to measure the similarity between two items; for more details, readers can refer to \cite{shi2019functional}. So for each target sentence, a collection of similar sentences can be found and ranked in descent order to select top N most similar ones for description extension.

Safovish \textit{et al.} \cite{safovish2020fiction} design a neural sentence expander trained on a corpus of fiction sentence compressions for the task of sentence expansion and enhancement, which is the state-of-the-art method on existing datasets (e.g., Fiction Corpus). In this method, a Seq2Seq model is used to predict sentences from the original input. To tackle the problem of copying input words during generation process, the modified negative log-likelihood loss function is used to increase the significance of learning new words. Specifically, the modified cross-entropy of generating is caculated as:
\begin{equation}
\mathcal{L} = -\sum_{t}^{}(1+\lambda \textrm{I}_{T-S}(w_{t}))\textrm{log}p(w_{t}|w_{1},...,w_{t-1}),
\end{equation}
in which I denotes the indicator function for a set, $T$ the ground truth, $S$ the source tokens, $\lambda$ is a parameter which controls the learning importance of the words in target sentence while not in original input. With this modification, the model no longer degenerates to
copying its input and can be trained as desired. In addition, to increase the novelty of generated sentence, Safovish \textit{et al.} \cite{safovish2020fiction} also develop a controlled sampling method for the decoding process so that the model can generate diverse output.

\subsubsection{Topic-to-essay Generation}
With the popularity of deep learning, the RNN-based Seq2Seq framework has been widely used in this field. Many works have been proposed to improve the traditional Seq2Seq framework.

To improve the global coherence of the generated essays, Feng \textit{et al.} \cite{feng2018topic-to-essay} propose an LSTM model with attention mechanism for essay generation. The main idea of this work is to maintain a topic coverage vector, each dimension of which represents the degree to which a topic word needs to be expressed in future generation, to adjust the attention policy, so that the model can consider more about unexpressed topic words. Specifically, the topic coverage vector is updated by parameter $\phi _{j}$, and it can be regarded as the discourse-level weight of $topic_{j}$, the word embedding of topic word $i$. Topic coverage vector $C_{t}$ is initialized as a k dimensional vector, i.e. $C_{0}$,  k is the number of input topic words, and each value of $C_{0}$ is 1.0. At time step $t$ in generation process, each element $c_{t,j}$ is updated as follows:
\begin{equation}
c_{t,j}=c_{t-1,j}-\frac{\alpha _{t,j}}{\phi _{j}},
\end{equation}
in which $\alpha _{t,j}=\textrm{exp}(g_{tj})/\sum_{i=1}^{K}\textrm{exp}(g_{ti})$ is the attention weight of topic word $i$ at time step $t$, and $g_{tj}$ is the attention score on $topic_{j}$ at time step $t$, which is caculated as:
\begin{equation}
g_{tj}=c_{t-1,j}v_{a}^{T}\textrm{tanh}(W_{a}h_{t-1}+U_{a}topic_{j}),
\end{equation}
where $h_{t-1}$ is the hidden representation of the LSTM at time step $t-1$, and $v_{a}$, $W_{a}$, $U_{a}$ are all parameters to be optimized. Therefore, the probability of the next word $y_{t}$ can be defined as:
\begin{equation}
\textrm{P}(y_{t}|y_{t-1},T_{t},C_{t})=\textrm{softmax}(\textrm{linear}(h_{t})),
\end{equation}
where the topic representation $T_{t}$ is formulated as $T_{t}=\sum_{i=1}^{K}\alpha _{tj}topic_{j}$.

Wang \textit{et al.} \cite{wang2020self-attention} propose an enhanced neural network based on self-attention and retrieval mechanisms, the encoder and decoder are constructed with self-attention to model longer dependence. And to alleviate the duplication problem, a retrieval process is adopted to collect topic-related sentences as an aid for essay generation. Specifically, input topic words are divided into $m$ groups, then the material $M=\left \{ S_{1},...,S_{m} \right \}$ can be collected based on cosine distance between the topic group and sentences in the corpus which is created by dividing the training set, where $S_{i}$ is a sentence that corresponds to the $i$-th topic group. For the different sequence relations between topic words and material sentences, two encoders with the same structure but different parameters are used to get the hidden representation $H^{Topic}=\left \{ h_{1}^{T},...,h_{k}^{T} \right \}$ and $H^{Material}=\left \{ h_{1}^{M},...,h_{ml}^{M} \right \}$, based on which the hidden states of generated of essay words $H=\left \{ h_{1},...,h_{n} \right \}$ are obtained by essay decoder. Finally the output probabilities of each essay word can be computed as:
\begin{equation}
p(y_{t}|y_{t-1},T,M)=\textrm{softmax}(\textrm{linear}(h_{t})).
\end{equation}

Meanwhile, Generative Adversarial Net (GAN) has shown promising results for topic-to-essay generation, which is effectively applied in \cite{yuan2019efficient} including a GAN model and two-level discriminators. The first discriminator guides the generator to learn the paragraph-level information and sentence syntactic structure with multiple LSTMs, and the second one processes higher level information such as topic and sentiment for essay generation. Then the reward is calculated based on results of two discriminators, and the generator $G_{\theta}$ tries to maximize expected reward from the initial state till the end state via the formulation:
\begin{gather}
J(\theta )=\sum_{t=1}^{T}E(R_{t}|S_{t-1},\theta )=\sum_{t=1}^{T}G_{\theta }(y_{t}|Y)[\lambda (D_{\phi }(Y))+(1-\lambda )D_{\gamma }(Y)],
\end{gather}
where $\lambda$ is a manually set weight, $Y$ is a complete sequence, $R_{t}$ is the reward for a whole sequence, $D_{\phi }$ and $D_{\gamma }$ are the first and second discriminator, respectively.

However, the generated essays of previous works are still lack of novelty and topic-consistency, based on which, some works leverage the external knowledge to solve this problem. Yang \textit{et al.} \cite{yang2019enhancing} introduce a model with the aid of commonsense knowledge in ConceptNet, in detail, each topic is used as a query to retrive $k$ neighboring concepts with pre-trained embeddings stored as commonsense knowledge in a memory matrix $\textrm{M}_{0}$, in the decoding phase, the generator $G_{\theta}$ refers to the memory matrix for text generation, the hidden state of the decoder at time-step $t$ is:
\begin{equation}
s_{t}=\textrm{LSTM}(s_{t-1},[e(y_{t-1});c_{t};m_{t}]),
\end{equation}
where [;] means the concatenation of vectors, $y_{t-1}$ is the word generated at time-step $t-1$. $c_{t}$ is the context vector that is computed by integrating the hidden representations of the input topic sequence, and $m_{t}$ is the memory vector extracted from $\textbf{M}_{t}$ based on the attention mechanism. As the generation progresses, the topic information that needs to be expressed keeps changing, which requires the memory matrix to be dynamically updated, so for each memory entry $\textbf{M}_{t}^{i}$ in $\textbf{M}_{t}$, a candidate update memory $\widetilde{\textbf{M}}_{t}^{i}$ is computed as:
\begin{equation}
\widetilde{\textbf{M}}_{t}^{i}=\textrm{tanh}(\textbf{U}_{1}\textbf{M}_{t}^{i}+\textbf{V}_{1}e(y_{t})),
\end{equation}
where $\textbf{U}_{1}$ and $\textbf{V}_{1}$ are trainable parameters. To determine how much the $i$-th memory entry should be updated, the adaptive gate mechanism is adopted:
\begin{equation}
g_{t}^{i}=\textrm{sigmoid}(\textbf{U}_{2}\textbf{M}_{t}^{i}+\textbf{V}_{2}e(y_{t})),
\end{equation}
where $\textbf{U}_{1}$ and $\textbf{V}_{1}$ are trainable parameters. $\textbf{M}_{t}^{i}$ is updated by:
\begin{equation}
\textbf{M}_{t+1}^{i}=(\textbf{1}-g_{t}^{i})\odot \textbf{M}_{t}^{i}+g_{t}^{i}\odot \widetilde{\textbf{M}}_{t}^{i},
\end{equation}
in which \textbf{1} refers to the vector with all elements 1 and $\odot$ denotes point-wise multiplication. Following their work, Qiao \textit{et al.} \cite{qiao2020a-sentiment-controllable} propose a topic-to-essay generator based on the conditional variational auto-encoder framework to control the sentiment, and introduces a topic graph attention mechanism to sufficiently use the structured semantic information which is ignored in \cite{yang2019enhancing}, so that the quality of generated essays is further improved and reaches the state-of-the-art level at present on the mainstream datasets (e.g., ZhiHu).

\section{Text Rewriting and Reasoning}\label{text_rewriting_and_reasoning}
\begin{table*}
\centering
\caption{Natrual language generation models for text rewriting and reasoning.}\label{tab: text rewriting}
\resizebox{\textwidth}{!}{
\begin{tabular}{l|l|l}
\toprule
\textbf{Task}& \textbf{Model}& \textbf{Description} \\

\midrule

\multirow{11}{*}{Text Style Transfer}& Spearean Modern Language Translator \cite{jhamtani2017shakespearizing}& RNN \\
~& ARAE \cite{zhao2018adversarially}& RNN + GAN + WAE \\
~& FM-GAN \cite{chen2018adversarial}& RNN + GAN \\
~& Unpaired Sentiment-to-Sentiment Translator \cite{xu2018unpaired}& RNN + RL + Unsupervised \\
~& Non-Offencive Language Translator \cite{santos2018fighting}& CNN + RNN + Unsupervised \\
~& BST \cite{prabhumoye2018style}& CNN + RNN + Unsupervised \\
~& DualRL \cite{luo2019a-dual}& RNN + RL + Unsupervised \\
~& Exploration-Evaluation-based Text Style Translator \cite{fu2018style}& RNN + Metrics Design \\
~& Evaluating Style Transfer for Text \cite{mir2019evaluating}& Embedding + CNN + RNN + Metrics Design \\
~& Error Margins Matter in Style Transfer Evaluation \cite{tikhonov2019style}& VAE + RNN + Metrics Design \\
~& PPVAE \cite{duan2020pre-train}& WAE + GAN \\

\midrule

\multirow{10}{*}{Dialogue Generation}& MARM \cite{zhou2017mechanism-aware}& RNN \\
~& GAN-AEL \cite{xu2017neural}& CNN + RNN + GAN \\
~& AEM \cite{luo2018an-auto-encoder}& RNN + Context-response Interaction \\
~& AGMN \cite{li2020attention}& RNN + KG + Retrieval + Context-response Interaction \\
~& Post-KS \cite{lian2019learning}& RNN + KB \\
~& KADG \cite{jiang2020knowledge}& Transformer + KB \\
~& $P^2$ BOT \cite{liu2020you-impress-me}& Transformer + Multi-task Learning + Persona \\
~& KnowledGPT \cite{wang2020knowledge}& Transformer + Knowledge Selection \\
~& PLATO \cite{bao2020plato}& Transformer + Multi-task Learning \\
~& TransferTransfo \cite{wolf2019transfertransfo}& Transformer + Multi-task Learning \\

\bottomrule
\end{tabular}}
\end{table*}

\subsection{Task}
The target of text rewriting and reasoning is to make a reversion of the text or apply reasoning methods to generate responses, which mainly contains two subtopics: text style transfer and dialogue generation. 

Text style transfer is the method to transform the attribute style of the sentence while preserving its attribute-independent content. Dialogue generation aims to automatically generate approximate answers to a series of given questions in a dialogue system. We show several classical methods for each subtask in Table~\ref{tab: text rewriting}.

\subsection{Data}
\subsubsection{Text Style Transfer}
The following four datasets are widely applied for the task of text style transfer. 

\paragraph{Yelp Review} The Yelp Review dataset is provided by the Yelp Dataset Challenge\footnote{https://www.yelp.com/dataset/}, which contains a large amount of business review texts. This dataset contains 1.43M, 10K, and 5K pairs for training, validation, and testing, respectively. 

\paragraph{Amazon Food Review} \cite{mcauley2013from} creates this dataset by crawling reviews from the Fine Foods category of Amazon. This dataset contains 367K, 10K, and 5K pairs for training, validation, and testing, respectively.

\paragraph{EMNLP2017 WMT News} \cite{guo2018long} picks the News section data from the EMNLP2017 WMT\footnote{http://statmt.org/wmt17/translation-task.html} Dataset, which is a large long-text corpus consists of 646,459 words and 397,726 sentences. After being preprocessed, this dataset contains 278,686 and 10,000 sentences for training and testing, respectively. 

\paragraph{GYAFC} The Grammarly’s Yahoo Answers Formality Corpus (GYAFC) \cite{rao2018dear} is a large corpus for formality stylistic transfer. The informal sentences are collected from Yahoo Answers\footnote{https://answers.yahoo.com/answer} with domains of Entertainment \& Music and Family \& Relationships, and the corresponding formal sentences are created using Amazon Mechanical Turk. This dataset totally contains 106,000 sentence pairs.

\subsubsection{Dialogue Generation}
There are four popular datasets for dialogue generation as shown below.

\paragraph{DailyDialog} The DailyDialog dataset \cite{li2017dailydialog} is a high-quality multi-turn dialog dataset containing daily conversations. The dialogues in the dataset is formally written by human with reasonable speaker turns, and often concentrate on a certain topic. Statistically, this dataset contains 13,118 multi-turn dialogues, nearly 8 average speaker turns per dialogue, and about 15 average tokens per utterance. 

\paragraph{UDC} The Ubuntu Dialogue Corpus (UDC) \cite{lowe2015ubuntu} is created based on the two-person conversations about Ubuntu-related problems in the Ubuntu chat logs\footnote{http://irclogs.ubuntu.com/} from 2004 to 2015. This dataset contains about 1 million multi-turn dialogues, over 7 million utterances, and 100 million words.

\paragraph{Persona-chat} The Persona-chat dataset \cite{zhang2018personalizing} is an engaging and personal chit-chat dialogue dataset collected by Amazon Mechanical Turk. Each of the paired crowdworkers condition their dialogue on a given provided profile. This dataset contains 164,356 utterances in total. 

\paragraph{Wizard-of-Wikipedia} The Wizard-of-Wikipedia dataset \cite{dinan2018wizard} is a large crowd-sourced collection for open-domain dialogue. Each of the paired speakers conduct open-ended chit-chat, and one of the speakers need to link the knowledge to sentences from existing Wikipedia articles connected to the topic. This dataset contains 22,311 dialogues with 201,999 turns.

\subsection{Method}
\subsubsection{Text Style Transfer}
In this field, the RNN-based Seq2Seq framework and deep latent variable model are broadly adopted. Jhamtani \textit{et al.} \cite{jhamtani2017shakespearizing} design a copy-enriched sequence-to-sequence model to transform text from modern English to Shakespearean English. The model first uses a fixed pre-trained embedding vector to represent each token, and uses a bidirectional LSTM to encode sentences. In this work, $\overrightarrow{LSTM_{enc}}$ and $\overleftarrow{LSTM_{enc}}$ represent the for-
ward and reverse encoder. $h^{\overrightarrow{enc}}_t$ represent hidden
state of encoder model at step t. The following equations describe the model:
\begin{equation}
    h^{\overrightarrow{enc}}_t=\overrightarrow{LSTM_{enc}}(h^{enc}_{t-1},E_{enc}(x_t)),
\end{equation}
\begin{equation}
    h^{\overleftarrow{enc}}_t=\overleftarrow{LSTM_{enc}}(h^{enc}_{t+1},E_{enc}(x_t)),
\end{equation}
\begin{equation}
    h^{enc}_{t}=h^{\overrightarrow{enc}}_t+h^{\overleftarrow{enc}}_t.
\end{equation}

In this work, only the forward and backward encoder states are added, and the standard connection is not used because it does not add additional parameters.Then a mixture model of RNN and pointer network are employed to transfer the text style. The pointer module provides location
based attention, and output probability distribution
due to pointer network module can be expressed as
follows:
\begin{equation}
    p^{PTR}_t(w)=\sum_{x_j=w}{(\beta_j)}.
\end{equation}

Zhao \textit{et al.} \cite{zhao2018adversarially} propose an adversarially regularized autoencoder framework to generalize the adversarial autoencoder, which combines a discrete autoencoder with a regularized latent representation of GAN, and can be further formalized by the Wasserstein autoencoder.The model is trained with coordinate descent across: (1)the encoder and decoder to minimize reconstruction, (2) the critic function to approximate the $W$ term, (3) the encoder
adversarially to the critic to minimize $W$:
\begin{equation}
 1) \min_{\phi,\psi}{L_{rec}(\phi,\psi)=E_{X\sim P_*}[-\log{p_{\psi}(x|enc_{\psi}(x))}]},
\end{equation}
\begin{equation}
 2)\max_{w\in W}{L_{cri}(w)=E_{X\sim P_*}[f_w(enc_{\psi}(x))]-E_{\widetilde{z}\sim P_z}[f_w(\widetilde{z})]},
\end{equation}
\begin{equation}
 3)\min_{\phi}{L_{enc}(\phi)=E_{X\sim P_*}[f_w(enc_{\psi}(x))]-E_{\widetilde{z}\sim P_z}[f_w(\widetilde{z})]}.
\end{equation}
Here $enc_{\psi}$ is a deterministic encoder function. $P_z$ is the the prior distribution, and $f_w(\widetilde{z})$ is the critic/discriminator.

Chen \textit{et al.} \cite{chen2018adversarial} utilize optimal transport to improve the ability of traditional GAN in processing descrete texts with the objective of feature-mover's distance. In this work, the feature-mover's distance(FMD) between two sets of sentence features is then defined as:
\begin{equation}
    D_{FMD}(P_f,P_{f^{'}})=\min_{T\geq 0}{\sum_{i=1}^{m}{\sum_{j=1}^{n}{T_{ij}\cdot c(f_i,f_{j^{'}})}}}=\min_{T\geq 0}{<T,C>},
\end{equation}
where $\sum_{j=1}^{n}{T_{ij}}=\frac{1}{m}$ and $\sum_{i=1}^{m}{T_{ij}}=\frac{1}{n}$ are the constraints, and $<,>$ represents the Frobenius dotproduct. In this work, the transport cost is defined as the cosine distance: $c(f_i,f_{j^{'}})=1-\frac{f_i^{T}f_j^{'}}{{\left\|f_i\right\|_2}{\left\|f_i^{'}\right\|_2}}$ and $C$ is the cost matrix.

However, the problem of lacking supervised parallel data has not been well studied by the above works. To tackle this problem, many unsupervised methods have been proposed. Xu \textit{et al.} \cite{xu2018unpaired} propose a cycled reinforcement learning approach through the cooperation between the neutralization and emotionalization modules. The neutralization module extracts the non-emotional part of the sentence with a single LSTM and a self-attention based sentiment classifier, and the emotionalization module generates emotional words and add them to the semantic content with a bi-decoder based encoder-decoder framework. 

Santos \textit{et al.} \cite{santos2018fighting} propose an unsupervised style transfer model to convert offensive language into non-offensive one. A single collaborative classifier is used to train the encoder-decoder network, and an attention mechanism with a cycle consistency loss is adopted to preserve the content.  Prabhumoye \textit{et al.} \cite{prabhumoye2018style} realize the style transfer of gender, political slant and sentiment through an unsupervised back-translation method. The transferring process is divided into two stages. The first stage learns the latent representation with back-translation of the input sentence through a language translation model, and the second stage adopts an adversarial generation technology to enable the output to match the desired style.The latent representation with back-translation of the input sentence can be described as:
\begin{equation}
    z=Encoder(X_f;\theta_E),
\end{equation}
where, $x_f$ is the sentence $x$ in language $f$. $\theta_E$ represent the parameters of the encoder of language $f$ $\rightarrow$ language $e$ translation system

Luo \textit{et al.} \cite{luo2019a-dual} reformulate the traditional unsupervised style transferring task as a one-step mapping problem, and propose a dual reinforcement learning framework to train the source-to-target and target-to-source mapping models. In this work, two reward methods that can evaluate style accuracy and content preservation separately are proposed. $R_s=P(s_y|y^{'};\psi)$ formulate the style classifier reward where $\psi$ is the parameter of the classifier and is fixed during
the training process, and $R_c=P(x|y^{'};\phi)$ represents the reward for preserving content. To encourage the model to improve both
the content preservation and the style accuracy, the final reward is the harmonic mean of the above two rewards:
\begin{equation}
R=(1+\beta^2)\frac{R_c\cdot R_s}{(\beta^2 \cdot R_c)+R_s },
\end{equation}
where $\beta$ is a harmonic weight aiming to control the trade-off between the two rewards.

Moreover, another challenge in this field is that there does not exist reliable evaluation metrics, which is neglected by previous works. To alleviate this issue, Fu \textit{et al.} \cite{fu2018style} propose two aspects of evaluation metrics to measure the transfer strength and content preservation of style transfer. The transfer strength aims to evaluate whether the style is transferred through an LSTM-sigmoid classifier. The style is defined in (30). This classifier is based on keras examples2. Transfer strength accuracy is defined as $\frac{N_{right}}{N_{total}}$, $N_{total}$is the number
of test data, and $N_{right}$is the number of correct case which
is transferred to target style. 
\begin{equation}
    l_{style}=\begin{cases}
    paper(positive)  &output\leq 0.5\\
    news(negative)   &output \geq 0.5\\ 
    \end{cases}.
\end{equation}

The content preservation is used to evaluate the similarity between source and target texts and is calculated by the embedding cosine distance. Content preservation rate is defined as cosine distance (31) between source sentence embedding $v_s$ and target sentence embedding $v_t$.
\begin{equation}
    score=\frac{v_s^{T}v_t}{\left\|v_s\right\|\cdot \left\|v_t\right\|}.
\end{equation}

Mir \textit{et al.} \cite{mir2019evaluating} specify three aspects of evaluation metrics including style transfer intensity, content preservation, and naturalness. The style transfer intensity is measured by Earth Mover's Distance, the content preservation is calculated by METEOR and embedding-based metrics, and the naturalness is obtained by an adversarial evaluation method. 

It is also instructive that Tikhonov \textit{et al.} \cite{tikhonov2019style} also point out the three significant problems encountered in the evaluation metrics of style transfer. These problems mainly illustrate that the measures of style accuracy and content preservation are often different in various style transfer tasks. Therefore, they propose to take BLEU between input and human-rewritten texts into consideration, so as to better measure the performance of style transfer models. Additonally, Duan \textit{et al.} \cite{duan2020pre-train} propose the Pre-train and Plug-in Variational Autoencoder (PPVAE), which is a model-agnostic framework towards flexible conditional text generation and consists of PretrainVAE and PluginVAE, where PretrainVAE aims to learn the original style of the sentence, and PluginVAE aims to learn the latent space of new style. The PPVAE achieves the state-of-the-art performance on the Yelp Reviews dataset.

\subsubsection{Dialogue Generation}
The RNN-based or GAN-based Seq2Seq model is widely leveraged to handle this task. Zhou \textit{et al.} \cite{zhou2017mechanism-aware} propose a mechanism-aware neural machine based on a probabilistic RNN-based Seq2Seq framework. The model first uses latent embeddings to represent the corresponding mechanisms, then an encoder-diverter-decoder framework is leveraged to generate mechanism-aware context.  In this study, there are $M$ latent mechanisms ${M_i}_{i=1}^{M}$ for response generation. Then, $p(y|x)$ can be expanded as follows:
\begin{equation}
      p(y|x)=\sum_{i=1}^{M}{p(y,m_i|x)}=\sum_{i=1}^{M}{p(m_i|x)p(y|m_i,x)},
\end{equation}
where $p(m_i|x)$  represents the probability of the mechanism $m_i$ conditioned on $x$. This probability actually measures the degree that $m_i$ can generate the response for x. The bigger of this value is, the more degree that the mechanism $m_i$ can be used to generate the responses for $x$. Additionally, $p(y|m_i,x)$  measures the probability that the response y is generated by the mechanism $m_i$ for $x$. With the modeling of $p(m_i|x)$ and $p(y|m_i,x)$ the objective of likelihood maximization, namely
\begin{equation}
    \sum_{(x,y)\in D^{c}}{\log{p(y|x)}}=\sum_{(x,y)\in D^{c}}{\log{\sum_{i=1}^{M}{p(m_i|x)p(y|m_i,x)}}},
\end{equation}
is used to learn the mechanism embeddings  ${M_i}_{i=1}^{M}$ and other model parameters.

Xu \textit{et al.} \cite{xu2017neural} introduce a GAN framework comprising a generator, a discriminator, and an approximate embedding layer to generate informative responses. The generator uses a Seq2Seq model with GRU to generate responses, and the discriminator uses a convolutional neural network to judge the difference between human responses and machine responses. In the approximate embedding layer, the overall word embedding approximation is computed as:
\begin{equation}
    \hat{e}_{w_i}=\sum_{j=1}^{V}{e_j \cdot \hbox{softmax}(W_p(h_i+Z_i)+b_p)j},
\end{equation}
where $w_p$ and $b_p$ are the weight and bias parameters of the word projection layer, respectively, and $h_i$ is the hidden representation of word $w_i$, from the decoding procedure of the generator $G$.

However, previous works ignore the semantic and utterance relationships between the context and response, and thus the obtained responses are not satisfying. To solve this problem, Luo \textit{et al.} \cite{luo2018an-auto-encoder} propose an auto-encoder matching model with a mapping module. In this model, two auto-encoders are leveraged to learn the semantic representations, and the mapping module is used to learn the utterance-level dependency between the context and response.
In this mapping module, for simplicity, there is only a simple feedforward network for implementation. The mapping module $M_{\gamma}$ transforms the source semantic representation $h$ to a new representation $t$. To be specific, we implement a multi-layer perceptron (MLP) $g(\cdot)$ for $M_{\gamma}$ and train it by minimizing the L2-norm loss $J_3(\gamma)$ of the transformed representation $t=g(h)$ and the semantic representation of target response $s$:
\begin{equation}
    J_3(\gamma)=\frac{1}{2}{\left\|{t-s}\right\|_2^{2}}.
\end{equation}

Li \textit{et al.} \cite{li2020attention} design a dual encoder model with an attention mechanism and a graph attention network. The attention mechanism is responsible of capturing the relationship between context and response, and the graph attention network is used to integrate the knowledge connections of domain words. The concept representation in the domain knowledge is constructed by a series of triples, $G(x) =\{T1, T2, . . . , Tn\}$ where $T_i$ has the same concept node $u$ but different neighbor concept $v$ and the graph representation of the concept $g(x)$ can be calculated by graph attention mechanism as:
\begin{equation}
    g(x)=\sum_{i=1}^{n}{\alpha T_i[u_i^{e};v_i^{e}]}, \text{ where } \alpha T_i=\frac{exp({\beta _{T_i}})}{\sum_{j=1}^{n}exp(\beta _{T_i})}, \ \beta _{T_i}=\hbox{ReLU}([(u_i^{e})^{T} W v_i^{e}]),
\end{equation}
in which $(u_i, r_i, v_i) = R_i\in G(x) $is the i-th triple in the dataset.

Some researchers try to improve the response quality by incorporating external knowledge bases, and propose many strategies to select appropriate knowledge. Lian \textit{et al.} \cite{lian2019learning} present a knowledge selection mechanism by separating the posterior distribution from the prior distribution. The distance between the posterior and prior distributions are minimized by the KL divergence during training, and during inference, the knowledges are selected and incorporated into the response based on the prior distribution.The Kullback-Leibler divergence loss (KLDivLoss), to measure the proximity between the prior distribution and the posterior distribution, which is defined as follows:
\begin{equation}
    L_{KL}(\theta)=\sum_{i=1}^{N}{p(k=k_i|x,y)\log{\frac{p(k=k_i|x,y)}{p(k=k_i|x)}}},
\end{equation}
where $\theta$ denotes the model parameters.

Jiang \textit{et al.} \cite{jiang2020knowledge} propose a knowledge augmented response generation model to improve the knowledge selection and incorporation. The model consists of a divergent knowledge selector and a knowledge aware decoder, where the selector conducts a one-hop subject reasoning over facts to reduce the subject gap in the knowledge selection, and the decoder is used to efficiently incorporate the selected fact. In addition, Liu \textit{et al.} \cite{liu2020you-impress-me} concern about the importance of conversational understanding for the high-quality chit-chat systems and propose the Persona Perception Bot (i.e., $P^2$ BOT). Different from other existing models, $P^2$ BOT focus on a important and previously overlooked concept, mutual persona perception, which is more appropriate to describe the process of information exchange that enables interlocutors to understand each other. The $P^2$ BOT is also the current state-of-the-art model on the Persona-chat dataset.

Recently, pre-trained language models (e.g., BERT) have shown significant improvements over traditional RNN-based methods in many NLP tasks and are also applied to this task. Wang \textit{et al.} \cite{wang2020knowledge} propose an encoder-decoder framework containing a BERT encoder and a transformer decoder. The encoder is used to learn semantic representations for both unstructured text and conversational history, and the decoder is leveraged to generate the dialogue response.In the encoder of this work,the input embedding is the sum of its token embedding, knowledge indicating embedding and position embedding:
\begin{equation}
    I(x_i)=E(x_i)+T(x_i)+P(x_i),
\end{equation}
where $E(x_i)$, $T(x_i)$, $P(x_i)$ are word embedding, knowledge indication embedding and position embedding, respectively. The input embeddings are then fed into BERT model to get the knowledge and dialogue history encoding representations.

Bao \textit{et al.} \cite{bao2020plato} design a pre-training framework with discrete latent variables. The pre-training tasks include repsonse generation and latent act recognition, which are jointly pre-trained through a unified network with shared parameters. Furthermore, inspired by the core idea of transfer learning, Wolf \textit{et al.} \cite{wolf2019transfertransfo} propose the TransferTransfo, which uses the paradigm of transfer learning to fine-tune the powerful transformer models. The specific fine-tuning tasks they select include: language modeling task, next utterance retrieval task, and generation task. Different fine-tuning tasks endow TransferTransfo with generalization performance for dialogue generation tasks in different scenarios.

\section{From Image to Text Generation}\label{from_image_to_text_generation}
\subsection{Task}
The image-based text generation aims at explaining or summarizing the visual concept of the given image, which mainly consists of three parts: image caption, video caption, and visual storytelling. The purpose of image captioning is to generate summaries from an image. Based on image captions, video caption aims to generate the summary of a series of images. Visual storytelling not only identifies the correlation between objects in a single picture but also gives the logical relationship between consecutive sequential images. It should be noted that the language generation component of VQA \cite{antol2015vqa,manmadhan2020visual,anderson2018bottom-up} model is relatively similar to that of image caption. There exists the main distinction that current VQA systems \cite{cao2019interpretable,gong2021cross,cao2021linguistically,do2021multiple,vu2020question,gong2022vqamix} are focused on reasoning process and mainly designed to choose answers from a given candidate answer set, which is not quite related to the natural language generation. Several popular methods for each subtask are shown in Table~\ref{tab: from image to text}.

\begin{table*}
\centering
\caption{From Image to Text Generation}\label{tab: from image to text}
\resizebox{\textwidth}{!}{
\begin{tabular}{l|l|l}
\toprule
\textbf{Task}& \textbf{Model}& \textbf{Description} \\ 
\midrule
\multirow{7}{*}{Image Caption}& Show and Tell \cite{vinyals2015show}& CNN + LSTM \\
~& BUTD \cite{anderson2018bottom-up}& Faster-RCNN + LSTM \\
~& Knowing When to Look \cite{lu2017knowing}& CNN + LSTM \\
~& Exploring Visual Relationship for Image Captioning \cite{yao2018exploring}& Faster-RCNN + LSTM + GCN \\
~& Self-Critical Sequence Training \cite{rennie2017self-critical}& CNN + LSTM + RL \\
~& Auto-Encoding Scene Graphs \cite{yang2019auto-encoding}& CNN + LSTM + GCN + RL \\ 
~& OFA \cite{wang2022unifying}& Transformer + Multimodal \\
\midrule
\multirow{10}{*}{Video Caption}& LRCN \cite{donahue2015long}& CNN + LSTM \\
~& S2VT \cite{venugopalan2016improving}& CNN + LSTM + Knowledge \\
~& Dense Caption Events \cite{wu2019densely}& CNN + LSTM + Daps \\
~& Masked Transformer \cite{zhou2018end-to-end}& CNN + TCN + Transformer \\
~& Hierarchical Reinforcement Learning \cite{wang2018a-reinforced}& CNN + LSTM + RL \\
~& Adversarial Inference \cite{park2019adversarial}& CNN + LSTM + Discriminator \\
~& VideoBERT Pretrain \cite{sun2019videobert}& CNN + BERT \\
~& ActBERT Pretrain \cite{zhu2020actbert}& CNN + BERT + Multi-task Learning \\
~& ClipBERT \cite{lei2021less}& CNN + BERT + Clip Sampling \\
~& UniViLM \cite{luo2020univilm}& Transformer + Multimodal \\
\midrule
\multirow{4}{*}{Visual Storytelling}& Informative Visual Storytelling \cite{lin2019informative}& CNN + GRU \\
~& Knowledgeable Storyteller \cite{yang2019knowledgeable}& CNN + GRU + Graph \\
~& Composite Reward \cite{hu2020what}& CNN + RNN + MLE \\ 
~& KAGS \cite{li2022knowledge-enriched}& CNN + RNN + KG \\
\bottomrule
\end{tabular}}
\end{table*}

\subsection{Data}
\subsubsection{Image Caption}
The literature review of the image caption datasets is shown below.

\paragraph{Flickr30k} \cite{plummer2015flickr30k} contains 31,783 images collected from Flickr. Most of these images depict humans performing various activities. Each image is paired with 5 crowd-sourced captions.

\paragraph{COCO} \cite{lin2014microsoft} is the largest image captioning dataset, containing 82,783, 40,504 and 40,775 images for training, validation and test respectively. This dataset is more challenging, since most images contain multiple objects in the context of complex scenes. Each image has 5 human annotated captions.

\paragraph{Visual Genome}\cite{krishna2017visualgenome} is composed of dense annotations of objects, attributes, and relationships within each image to learn these models. Specifically, this dataset contains over 108K images where each image has an average of 35 objects, 26 attributes, and 21 pairwise relationships between objects.

\subsubsection{Video Caption}
There are mainly three popular datasets for video caption as shown below.

\paragraph{MSR-VTT} \cite{xu2016msrvtt} is the most widely used video caption dataset, which contains 7,180 videos of 20 categories. This is created by collecting 41.2 hours of 10K web video clips from a commercial video search engine.

\paragraph{Charades} \cite{sigurdsson2016charades} is collected with a focus on common household activities using the Hollywood in Homes approach. This dataset contains 9,848 videos with 66,500 annotations describing 157 actions. 

\paragraph{ActivityNet} \cite{krishna2017activitynet} is a large dataset that connects  videos to a series of temporally annotated sentences. Each sentence describes what occurs in an unique segment of a video. Specifically, AcitvityNet contains 20k videos with 100k sentence-level description.

\subsubsection{Visual Storytelling}
The following two datasets are widely used for visual storytelling.

\paragraph{VIST} \cite{huang2016visual} is the most widely used dataset for visual storytelling. It contains 10,032 visual albums with 50,136 stories. Each story contains five narrative sentences, corresponding to five grounded images respectively.

\paragraph{VideoStory} \cite{gella2018dataset} contains 20k videos posted publicly on a social media platform amounting to 396 hours of video with 123k sentences.

\subsection{Method}
\subsubsection{Image Captioning}
Image captioning aims to generate a description of the given image. The Show-Tell model \cite{vinyals2015show} proposed an encoder-decoder based framework that encodes images into feature vectors with Convolution Neural Networks (CNN), and decodes the feature vectors into words with Recurrent Neural Networks (RNN). To obtain the fine-grained visual concepts, attention-based image captioning model \cite{lu2017knowing} was proposed to ground words with the corresponding part of imaging. Considering the fact that region aware feature better fits the human visual system, Anderson \textit{et al.} \cite{anderson2018bottom-up} Propose a recognized baseline called Bottom-Up-Top-Down (BUTD) for image caption. The BUTD is composed of two LSTM layers. The first LSTM layer is designed to capture the top-down visual attention model, while the second LSTM layer is regarded as a language model. Given the mean-pooled image feature $\overline{\boldsymbol{v}}$, a word embedding matrix $W_{e}$, and a one-hot encoding $\Pi_{t}$ of the input word at time $t$ we could obtain the output $\boldsymbol{h}_{t}^{1}$ as:
\begin{equation}
\boldsymbol{h}_{t}^{1}=\mbox{LSTM} \left( \left[\boldsymbol{h}_{t-1}^{2}, \overline{\boldsymbol{v}}, W_{e} \Pi_{t}\right], \boldsymbol{h}_{t-1}^{1} \right),
\end{equation}
and the normalized attention weight $a_{i, t}$ can be represented by the following formulations:
\begin{align}
a_{i, t} &=\boldsymbol{w}_{a}^{T} \tanh \left(W_{v a} \boldsymbol{v}_{i}+W_{h a} \boldsymbol{h}_{t}^{1}\right), \\
\boldsymbol{\alpha}_{t} &=\operatorname{softmax}\left(\boldsymbol{a}_{t}\right),
\end{align}
The attended image feature used as input to the language LSTM is calculated as a convex combination of all input features as $\hat{v}_{t}=\sum_{i=1}^{K} \alpha_{i, t} v_{i}$. The input to the language model LSTM consists of the attended image feature, concatenated with the output of the attention LSTM, given by $\boldsymbol{h}_{t}^{2}=\left[\hat{\boldsymbol{v}}_{t}, \boldsymbol{h}_{t}^{1}\right]$. Using the notation $y_{1:T}$ to refer to a sequence of words, at each time step t the conditional distribution
over possible output words is given by:
\begin{equation}
p\left(y_{t} \mid y_{1: t-1}\right)=\operatorname{softmax}\left(W_{p} \boldsymbol{h}_{t}^{2}+\boldsymbol{b}_{p}\right),
\end{equation}
The distribution over complete output sequences is calculated as the product of conditional distributions: 
\begin{equation}
p\left(y_{1: T}\right)=\prod_{t=1}^{T} p\left(y_{t} \mid y_{1: t-1}\right).
\end{equation}

To reduce exposure bias and metric mismatching in sequential training, notable efforts are made to optimise non-differentiable metrics using reinforcement learning \cite{liu2017improved, rennie2017self-critical, xu2020multi-level}. To further boost accuracy, detected semantic concepts \cite{gan2017semantic, wu2016what, you2016image} are adopted in captioning framework. A more structured representation over concepts calling scene graph is further explored \cite{yang2019auto-encoding, yao2018exploring} in image captioning which can take advantage of detected objects and their relationships. Instead of using a fully detected scene graph to improve captioning accuracy, Chen \textit{et al.} \cite{chen2020say} propose to employ Abstract Scene Graph as control signal to generate intention-aware and diverse image captions. 

With the progress of multimodal representation learning, Wang \textit{et al.} \cite{wang2022unifying} propose the OFA, a unified multimodal pretrained model that can be applied in all modalities and various tasks. The OFA is simple yet effective, and it achieves new state-of-the-art performance on the kinds of multimodal tasks, such as image captioning, text-to-image generation, and VQA.

\subsubsection{Video Caption}
The aim of video caption is to describe or summarize a video in natural language. It is a non-trivial task for computers since it is difficult to select the useful visual features from a video clip and describe what's happening in a way that obeys the common sense of humanity. 

The currently prevailing architecture for video caption is composed of a CNN-like visual encoder and a RNN-like linguistic decoder. Donahue \textit{et al.} \cite{donahue2015long} design the Long-term Recurrent Convectional Networks (LRCNs) that is both temporally and spatially deep. After that, Venugopalan \textit{et al.} \cite{venugopalan2016improving} introduce the S2VT, a Seq2Seq approach for video to text with the knowledge from text corpora. Krishna \textit{et al.} \cite{krishna2017activitynet} propose a captioning module that uses contextual information from past and future events to capture the dependencies between the events in a video. However, Krishna \textit{et al.} \cite{krishna2017activitynet} fails to take advantage of language to benefit event proposal with the co-training diagram. Thus, Zhou \textit{et al.} \cite{zhou2018end-to-end} propose a video caption framework that produces proposal and description simultaneously. To describe a video with multiple fine-grained actions, Wang \textit{et al.} \cite{wang2018a-reinforced} propose a hierarchical reinforcement learning framework that a high-level agent learns to design sub-goals and a low-level worker recognizes the primitive actions to fulfill the sub-goal. By introducing the discriminator that considers visual relevance to the video, language diversity \& fluency, and coherence across sentences, Park \textit{et al.} \cite{park2019adversarial} generate more accuracy video descriptions. To resolve the dilemma that encoders of vision-language tasks are not trained end to end, Lei \textit{et al.} \cite{lei2021less} propose ClipBERT, a framework that applies the sparse sampling to use a few sampled clips to achieves better performance. Additionally, with the rise of multimodal learning, Luo \textit{et al.} \cite{luo2020univilm} propose UniVL, a unified multimodal pre-training model for vedio captioning. The UniVL consists of four components (i.e., two encoders for single-modal, a cross-modal encoder and a decoder) and is pre-trained with five tasks, including language understanding and generation tasks. The highlight of UniVL is that it uses both understanding and generative tasks for cross-modal pre-training, leading to its state-of-the-art performance on video captioning.

Recent years, self-supervised learning has become increasingly important with its power to leverage the abundance of unlabeled data. Sun \textit{et al.} \cite{sun2019videobert} propose VideoBERT to learn bidirectional joint distributions over sequences of visual and linguistic tokens without any explicit supervision. Zhu \textit{et al.} \cite{zhu2020actbert} propose ActNet, which models global and local visual cues for fine-grained visual and linguistic relation learning.

\subsubsection{Visual Storytelling}
Visual Storytelling not only needs to identify the correlation between objects in a single picture, but also Need to identify and learn the logical relationship between consecutive sequential images. In practice, Visual Storytelling is prone to problems such as single narrative words, rigid sentences, incoherent context logic, and lack of emotion in story descriptions. aims at generating a coherent and reasonable story with a series of images \cite{huang2016visual}. To deal with the issue that Visual Storytelling usually focus on generating general description rather than the details of meaningful visual contents, Li \textit{et al.} \cite{lin2019informative} propose to mine the cross-modal rules to assistant the concept inference. Yang \textit{et al.} \cite{yang2019knowledgeable} present a commonsense-driven generative model to introduce crucial commonsense from the external knowledge base for visual storytelling. Due to the limitation of maximum likelihood estimation on training, the majority of existing models encourage high resemblance to texts in the training database, which makes the description overly rigid and lack in diverse expressions. Therefore, Mo \textit{et al.} \cite{mo2019adversarial} cast the task as a reinforcement learning task and propose an Adversarial All-in-one Learning (AAL) framework to learn a reward model, which simultaneously incorporates the information of all images in the photo stream and all texts in the paragraph, and optimize a generative model with the estimated reward. To make the Visual Storytelling model topic adaptively, Li \textit{et al.} \cite{lin2019informative} introduce a gradient-based meta-learning algorithm. Conventional storytelling approaches usually focused on optimizing metrics such as BLEU, ROUGE and CIDEr. In this paper, Hu \textit{et al.} \cite{hu2020what} revisit the issue from a different perspective, by delving into what defines a natural and thematically coherent story. In addition, considering the inability of previous methods to explore latent information beyond the image and thus fail to capture consistent dependencies from the global representation, Li \textit{et al.} \cite{li2022knowledge-enriched} propose the KAGS, a knowledge-enriched attention network with group-wise semantic model which achieves new state-of-the-art performance with respect to both objective and subjective evaluation metrics.

\section{NLG Evaluation Metrics}\label{evaluation_metrics}
In the research field of Artificial Intelligence, the evaluation metrics for models of kinds of tasks have always been the focus of attention for a long time, and the same is true in the fields of NLP \cite{su1992a-new-quantitative}.
In this section, we mainly introduce several automatic evaluation metrics for NLG, which can be divided into two categories: untrained evaluation metrics, and machine-learned evaluation metrics \cite{celikyilmaz2021evaluation}.

\subsection{Untrained Evaluation Metrics}
This category of metric is most widely used in the NLG community since it is easy to be implemented and does not involve additional training cost, which compares machine-generated texts to human-generated ones simply based on content overlap, string distance or lexical diversity. We mainly introduce five metrics of such category, including BLEU, ROUGE, METEOR, Distinct, and Self-BLEU. 

The Bilingual Evaluation Understudy (BLEU) metric \cite{papineni2002bleu} is used to calculate the co-occurrence frequency of two sentences based on the weighted average of matched n-gram phrases. BLEU was originally used to evaluate machine translation, and has been used for more and more NLG tasks, such as question generation \cite{zhao2018paragraph-level}, topic-to-essay generation \cite{yang2019enhancing}, text style transfer \cite{luo2019a-dual}, and dialogue generation \cite{bao2020plato,lin2021graph}.

The Recall-Oriented Understudy for Gisting Evaluation (ROUGE) metric \cite{lin2004rouge} is used to measure the similarity between the generated and reference texts based on the recall score. This metric is commonly used in the field of text summarization, including four types: ROUGE-n measures the n-gram co-occurrence statistics; ROUGE-l measures the longest common subsequence; ROUGE-w measures the weighted longest common subsequence; ROUGE-s measures the skip-bigram co-occurrence statistics. ROUGE has also been widely applied to other NLG tasks such as question generation \cite{zhao2018paragraph-level}, distractor generation \cite{qiu2020automatic}, and dialogue generation \cite{bao2020plato}.

The Metric for Evaluation of Translation with Explicit Ordering (METEOR) metric \cite{banerjee2005meteor} is an improvement over BLEU to address several weaknesses including four aspects: lack of recall, use of higher order n-grams, lack of explicit word-matching between translation and reference, and use of geometric averaging of n-grams, which is calculated by the harmonic mean of the unigram precision and recall. In addition to machine translation, METEOR has also been widely used in text summarization \cite{qi2020prophetnet}, question generation \cite{zhao2018paragraph-level}, and dialogue generation \cite{bao2020plato}.

The Distinct metric \cite{li2016diversity} is used to measure the diversity of response sequences for dialogue generation. It calculates the number of distinct unigrams and bigrams in generated responses to reflect the diversity degree. To avoid preference for long sequences, the value is scaled by the total number of generated tokens. 

The Self-BLEU metric \cite{zhu2018texygen} is also a metric to measure the diversity. Different from BLEU that only evaluates the similarity between two sentences, Self-BLEU is used to measure the resemblance degree between one sentence (hypothesis) and the rest sentences (reference) in a generated collection. It first calculates the BLEU score of every generated sentence against other sentences, then the average BLEU score is defined as the Self-BLEU score of the document, where a lower Self-BLEU score implies higher diversity.

\subsection{Machine-learned Evaluation Metrics}
This category of metric is based on machine-learned models to simulate human judges, which evaluates the similarity between machine-generated texts or between machine-generated texts and human-generated ones. We mainly introduce three metrics of such category, containing ADEM, BLEURT, and BERTScore. 

The Automatic Dialogue Evaluation Model (ADEM) metric \cite{lowe2017towards} is used to automatically evaluate the quality of dialogue responses, where the evaluation model is trained in a semi-supervised manner with a hierarchical recurrent neural network (RNN) to predict the response scores. Specifically, given the dialogue context $\mathbf{c}$, model response $\mathbf{\hat{r}}$, and reference response $\mathbf{r}$ encoded by a hierarchical RNN, the predicted score can be calculated by:
\begin{equation}
score = (\mathbf{c}^{\top} M \mathbf{\hat{r}} + \mathbf{r}^{\top} N \mathbf{\hat{r}} - \alpha) / \beta,
\end{equation}
where $M, N$ are learnable matrices initialized by the identity, and $\alpha, \beta$ are scalar constants to initialize the predicted scores in range [1,5].

The Bilingual Evaluation Understudy with Representations from Transformers (BLEURT) metric \cite{sellam2020bleurt} is based on BERT \cite{devlin2019bert} with a novel pre-training scheme. Before fine-tuning BERT on rating data to predict human rating scores, a pre-training method is applied, where BERT is pre-trained on a large number of synthetic sentence pairs on several lexical- and semantic-level supervision signals in a multi-task manner. This pre-training process is important and can improve the robustness to quality drifts of generation systems. 

The BERTScore metric \cite{zhang2020bertscore} uses pre-trained contextual embeddings from BERT to measure the similarity between two sentences. Given the contextual embeddings of a reference sentence $x$ and a candidate sentence $\hat{x}$, namely $\mathbf{x}, \mathbf{\hat{x}}$, the recall, precision, and F1 scores are calculated by:
\begin{align}
R_{\rm BERT} &= \frac{1}{|x|} \sum_{x_i \in x} \max_{\hat{x}_j \in \hat{x}} \mathbf{x}_i^{\top} \mathbf{\hat{x}}_j, \\
P_{\rm BERT} &= \frac{1}{|\hat{x}|} \sum_{\hat{x}_j \in \hat{x}} \max_{x_i \in x} \mathbf{x}_i^{\top} \mathbf{\hat{x}}_j, \\
F_{\rm BERT} &= 2 \frac{P_{\rm BERT} \cdot R_{\rm BERT}}{P_{\rm BERT} + R_{\rm BERT}},
\end{align}
where the recall is calculated by matching each token in $x$ to a token in $\hat{x}$, the precision is obtained by matching each token in $\hat{x}$ to a token $x$, and greedy matching is adopted to match the most similar tokens.

\subsection{Human Evaluation Metrics}
For generation, human evaluation focuses on the explanation of two key matters: diversity and creativity, i.e., the capacity of varying their texts in form and emphasis to fit an enormous range of speaking situations, and the potential to express any object or relation as a natural language text.
In further detail, human evaluation is implemented to evaluate on three aspects: Grammar (whether a generated sentence is fluent without grammatical error), Faithful (whether the output is faithful to input), and Coherent (whether a sentence is logically coherent and the order of expression is in line with human writing habits). This needs to organize the capabilities of the people who work on generation in field of computational linguistics and artificial intelligence.

\section{Problems And Challenges}\label{problems_and_challenges}
In this section, we primarily point out four problems and challenges that deserve to be tackled and investigated further, including the evaluation method, external knowledge engagement, controllable generation, and multimodal scenarios.

\paragraph{Evaluation Method} 
Evaluation method is still an important and open research area for the field of NLG. As pointed by \cite{deriu2021survey}, traditional untrained evaluation metrics do not always correlate well with human judgements, while recent machine-learned metrics need a large amount of human annotations and not always have good transferability. Hence, there still exists a significant amount of challenges and improvement room in this area.

\paragraph{External Knowledge Engagement} 
Considering the limited information lying in the original texts and the difficulty of generating satisfying sentences \cite{yu2021survey}, it is crucial to incorporate external knowledge to enhance the performance. Therefore, how to obtain useful and correlative knowledge and how to effectively incorporate the knowledge still deserve to be investigated.

\paragraph{Controllable Generation} 
Another challenging problem is how to generate controllable natural language as we would like it to be. Although a great body of work has been done in this area to study how to perform various kinds of controlled text generation, there is still a lack of uniform paradigms and standards about it. More importantly, how to measure the controllability of the generated text remains an open question, for different controlled contents.

\paragraph{Multimodal Scenarios} 
Recently, research on various applications in multimodal scenarios have gradually attracted more and more attention from NLP researchers. How to apply natural language generation methods in multimodal scenarios has been a worthy problem and promising direction. It is reasonable to believe that the utilization of rich multimodal information into natural language generation tasks will surely further advance the progress and development in this direction.

\section{Conclusions}
\label{conclusions}
Over the past few years, natural language generation tasks and methods have become important and indispensable in natural language processing. This progress owes to advances in various deep learning-based methods. This article describes deep learning research on natural language generation with a historical perspective, emphasizing the special character of the problems to be solved. It begins by contrasting generation with language understanding, establishing basic concepts about the tasks, datasets, and the deep learning methods through it. A section of evaluation metrics from the output of generation systems follows, showing what kinds of performance are possible and where the difficulties are. Finally, some open problems are suggested to indicate the major challenges and future research directions of natural language generation.

\section*{Acknowledgement}
This work was supported in part by the 173 program No. 2021-JCJQ-JJ-0029, the Shenzhen General Research Project under Grant JCYJ20190808182805919 and in part by the National Natural Science Foundation of China under Grant 61602013.

\bibliographystyle{ACM-Reference-Format}
\bibliography{csur}

\end{document}